\documentclass[10pt,logo,copyright]{nvidiatechreport}
\usepackage[authoryear,sort&compress,round]{natbib}

\usepackage[utf8]{inputenc} 
\usepackage[T1]{fontenc}    

\usepackage{parskip}        
\usepackage{url}            
\usepackage{booktabs}       
\usepackage{amsfonts}       
\usepackage{nicefrac}       
\usepackage{microtype}      
\usepackage{xcolor}         
\usepackage[dvipsnames]{xcolor} 
\usepackage{graphicx}
\usepackage{animate}        
\usepackage{subcaption}
\usepackage{tabularx}
\usepackage{makecell}
\usepackage{tcolorbox}
\usepackage{fancyvrb}
\usepackage{fvextra}
\usepackage{adjustbox}
\usepackage{setspace}
\newcolumntype{M}[1]{>{\centering\arraybackslash}m{#1}}
\usepackage{float}
\usepackage{tikz}
\usetikzlibrary{positioning,shapes,arrows}
\usepackage{amsmath,amsfonts,bm, bbm,leftindex}
\usepackage{multirow}
\usepackage{comment}
\usepackage{gensymb}
\usepackage{lipsum}
\usetikzlibrary{arrows.meta, positioning, fit}
\usepackage[para]{threeparttable}
\usepackage{tikz}
\usetikzlibrary{tikzmark}

\def\eqref#1{equation~\ref{#1}}

\def\1{\bm{1}}

\DeclareMathAlphabet{\mathsfit}{\encodingdefault}{\sfdefault}{m}{sl}
\SetMathAlphabet{\mathsfit}{bold}{\encodingdefault}{\sfdefault}{bx}{n}

\makeatletter
\let\save@mathaccent\mathaccent
\newcommand*\if@single[3]{%
  \setbox0\hbox{${\mathaccent"0362{#1}}^H$}%
  \setbox2\hbox{${\mathaccent"0362{\kern0pt#1}}^H$}%
  \ifdim\ht0=\ht2 #3\else #2\fi
  }
\newcommand*\rel@kern[1]{\kern#1\dimexpr\macc@kerna}
\newcommand*\widebar[1]{\@ifnextchar^{{\wide@bar{#1}{0}}}{\wide@bar{#1}{1}}}
\newcommand*\wide@bar[2]{\if@single{#1}{\wide@bar@{#1}{#2}{1}}{\wide@bar@{#1}{#2}{2}}}
\newcommand*\wide@bar@[3]{%
  \begingroup
  \def\mathaccent##1##2{%
    \let\mathaccent\save@mathaccent
    \if#32 \let\macc@nucleus\first@char \fi
    \setbox\z@\hbox{$\macc@style{\macc@nucleus}_{}$}%
    \setbox\tw@\hbox{$\macc@style{\macc@nucleus}{}_{}$}%
    \dimen@\wd\tw@
    \advance\dimen@-\wd\z@
    \divide\dimen@ 3
    \@tempdima\wd\tw@
    \advance\@tempdima-\scriptspace
    \divide\@tempdima 10
    \advance\dimen@-\@tempdima
    \ifdim\dimen@>\z@ \dimen@0pt\fi
    \rel@kern{0.6}\kern-\dimen@
    \if#31
      \overline{\rel@kern{-0.6}\kern\dimen@\macc@nucleus\rel@kern{0.4}\kern\dimen@}%
      \advance\dimen@0.4\dimexpr\macc@kerna
      \let\final@kern#2%
      \ifdim\dimen@<\z@ \let\final@kern1\fi
      \if\final@kern1 \kern-\dimen@\fi
    \else
      \overline{\rel@kern{-0.6}\kern\dimen@#1}%
    \fi
  }%
  \macc@depth\@ne
  \let\math@bgroup\@empty \let\math@egroup\macc@set@skewchar
  \mathsurround\z@ \frozen@everymath{\mathgroup\macc@group\relax}%
  \macc@set@skewchar\relax
  \let\mathaccentV\macc@nested@a
  \if#31
    \macc@nested@a\relax111{#1}%
  \else
    \def\gobble@till@marker##1\endmarker{}%
    \futurelet\first@char\gobble@till@marker#1\endmarker
    \ifcat\noexpand\first@char A\else
      \def\first@char{}%
    \fi
    \macc@nested@a\relax111{\first@char}%
  \fi
  \endgroup
}
\makeatother

\usepackage{amsmath}
\usepackage{amssymb}
\usepackage{epsfig}
\usepackage{lipsum}
\usepackage{amsthm}
\usepackage{comment}
\usepackage{multirow}
\usepackage{subcaption}
\usepackage{xspace}
\usepackage{setspace}
\usepackage[utf8]{inputenc} 
\usepackage[T1]{fontenc}    
\usepackage{enumitem}
\usepackage{makecell}
\usepackage{gensymb}
\usepackage{array}
\usepackage{nimbusmononarrow}

\usepackage{parskip}        
\usepackage{url}            
\usepackage{booktabs}       
\usepackage{amsfonts}       
\usepackage{nicefrac}       
\usepackage{microtype}      
\usepackage{booktabs}
\usepackage{multirow}
 
\newcommand{\sevenbupscaler}{Cosmos-Transfer1-7B-4KUpscaler\xspace}
\newcommand{\sevenbav}{Cosmos-Transfer1-7B-Sample-AV\xspace}
\newcommand{\sevenb}{Cosmos-Transfer1-7B\xspace}
\newcommand{\modelname}{Cosmos-Transfer1\xspace}
\newcommand{\evalsetname}{TransferBench\xspace}

\usepackage[nameinlink]{cleveref}
\crefname{equation}{Eq.}{Eqs.}
\crefname{figure}{Fig.}{Figs.}
\crefname{section}{Sec.}{Sec.}
\crefname{appendix}{App.}{App.}
\crefname{table}{Tab.}{Tabs.}
\crefname{algorithm}{Algo}{Algo}
\crefname{thm}{Thm}{Thm}
\Crefname{thm}{Thm}{Thm}
\crefname{prop}{Prop}{Prop}

\newcommand{\crefnames}[3]{%
  \@for\next:=#1\do{%
    \expandafter\crefname\expandafter{\next}{#2}{#3}%
  }%
}

\title{\modelname: Conditional World Generation with Adaptive Multimodal Control}

\author{
    NVIDIA\footnote{A detailed list of contributors and acknowledgments can be found in~\cref{sec:contributors} of this paper.}
}

\begin{abstract}
We introduce \modelname, a conditional world generation model that can generate world simulations based on multiple spatial control inputs of various modalities such as segmentation, depth, and edge. In the design, the spatial conditional scheme is adaptive and customizable. It allows weighting different conditional inputs differently at different spatial locations. This enables highly controllable world generation and finds use in various world-to-world transfer use cases, including Sim2Real. We conduct extensive evaluations to analyze the proposed model and demonstrate its applications for Physical AI, including robotics Sim2Real and autonomous vehicle data enrichment. We further demonstrate an inference scaling strategy to achieve real-time world generation with an NVIDIA GB200 NVL72 rack. To help accelerate research development in the field, we open-source our models and code at \url{https://github.com/nvidia-cosmos/cosmos-transfer1}.
\end{abstract}

\begin{document}
\maketitle
\abscontent
\section{Introduction}\label{sec::intro}

Multimodal controllable world generation refers to the problem of generating world simulation videos based on multimodal video inputs such as segmentation, depth, and edge. These inputs help define specifics of the target world at different spatial locations at different time instances, which simplifies the generation problem. Such a controllable generation capability is valuable. One can leverage the capability to mitigate the synthetic-to-real domain gap problem of a CG-based simulator. We can make the renderings more realistic while preserving the scene structure and semantic through using depth and segmentation, often freely available in the CG-based simulator, as the multimodal inputs in world generation.

In this paper, we propose \modelname, a diffusion-based conditional world model for the multimodal controllable world generation problem. Our model is built on top of Cosmos-Predict1~\citep{nvidia2025cosmos}, which consists of a set of diffusion transformer-based (DiT-based)~\citep{peebles2023scalable} world models. We add multimodal control branches to the DiT through a novel ControlNet design~\citep{zhang2023adding}. We build a control branch per modality. If there are $N$ multimodal video inputs, then we will have $N$ control branches. We train the $N$ control branches separately and fuse them in the inference time.

Our multimodal control is spatially and temporally adaptive. This is achieved by applying a spatiotemporal control map to the outputs of the control branches. The spatiotemporal control map specifies the weight for each modality at each location and time instance. The higher the weight, the more influence the modality has on the generation output at the location and the time instance. This adaptive weighting scheme enables various ways to control the generation output. For example, one can choose to favor depth modality input more across all the locations in order to favor preservation of the input scene geometry. One can also give edge modality more weight to the foreground object to preserve the fine-grained details of the foreground object and less weight to the background to allow diverse background generation. The spatiotemporal control map could also be inferred by a separate neural module.

We conduct extensive empirical evaluations to verify the effectiveness of \modelname. We measure its generation quality and controllability on several Physical AI related world generation tasks. We also discuss applications of \modelname for robotics Sim2Real and autonomous vehicle data enrichment. We further demonstrate an inference scaling strategy to achieve real-time world generation with an NVIDIA GB200 NVL72 rack. To help advance the field, our code, model weights, and example scripts are open-sourced at \url{https://github.com/nvidia-cosmos/cosmos-transfer1}.

\section{Preliminary}\label{sec::preliminary}

\begin{figure*}[t!]
\centering
    \includegraphics[width=0.99\textwidth]{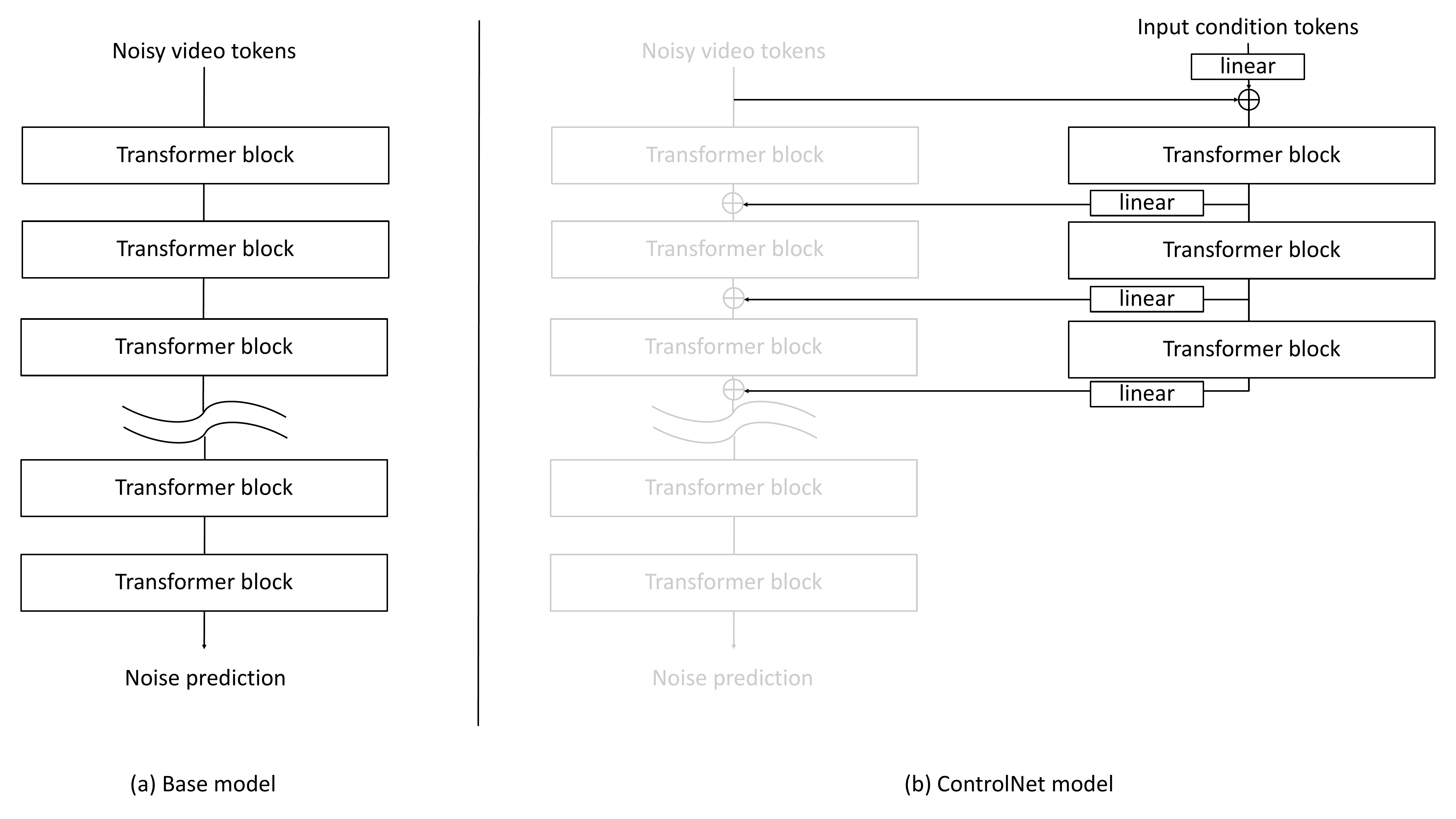}
    \caption{(a) \textbf{Base model} is the base DiT-based diffusion model. It consists of a sequence of transformer blocks and learns to predict the added noise in the input noisy tokens. (b) \textbf{ControlNet} extends the base model to a conditional diffusion model. The main addition is the control branch, which contains a few transformer blocks. The outputs of the transformer blocks are passed to zero-initialized linear layers before added back to the main branch. During the ControlNet training, the base model weights are frozen.}
    \label{fig:basecontrolnet}
\end{figure*}

The main component of a diffusion model is a denoiser. A common approach to implement the denoiser is to use the DiT architecture as visualized in~\cref{fig:basecontrolnet}(a). The architecture consists of a sequence of transformer blocks, trained to predict the noise $\mathbf{b}$ added to the input video tokens. Let $D$ denote the denoiser. In the diffusion model, it takes noisy video tokens $\mathbf{x}_{\sigma}$ and noise deviation $\sigma$ as inputs and uses them to predict the noise. That is 
\begin{equation}
    \mathbf{n} = D(\mathbf{x}_{\sigma}, \sigma).
\end{equation}

ControlNet is a popular approach to extend the base diffusion model to a conditional diffusion model. The original ControlNet was developed for the UNet model. \cite{chen2024pixart} extended it to the DiT architecture. Based on the prior works, we design our DiT-based ControlNet visualized in~\cref{fig:basecontrolnet}(b). It has the base model and a control branch. The base model is the base diffusion model. The control branch consists of a few transformer blocks for conditional inputs. These conditional inputs serve as the control signal. In our ControlNet design, we use three conditional blocks. We choose three as it empirically offers a good balance between control effectiveness and inference efficiency. These transformer blocks are initialized by inheriting the weights from the corresponding transformer blocks in the base diffusion model. Outputs from these blocks are passed to corresponding linear layers, respectively. These linear layers are zero-initialized. The linear layer outputs are then added to the activations of the corresponding transformer block in the base model. During training, these additional blocks are optimized while the base model is kept frozen. A well-trained denoiser learns to leverage information in the conditional tokens to predict the noise. Let $\mathbf{c}$ be the conditional tokens. The denoiser for the ControlNet becomes a conditional denoiser. It predicts $\mathbf{n}$ based on $\mathbf{x}_{\sigma}$, $\sigma$, and $\mathbf{c}$.
\begin{equation}
    \mathbf{n} = D(\mathbf{x}_{\sigma}, \sigma, \mathbf{c}).
\end{equation}
We also note that ControlNet inherits the base model capability. When the base diffusion model takes text prompts as inputs, the ControlNet also takes text prompts as inputs.
\section{Method}\label{sec::method}

\begin{figure*}[t!]
\centering
    \includegraphics[width=0.99\textwidth]{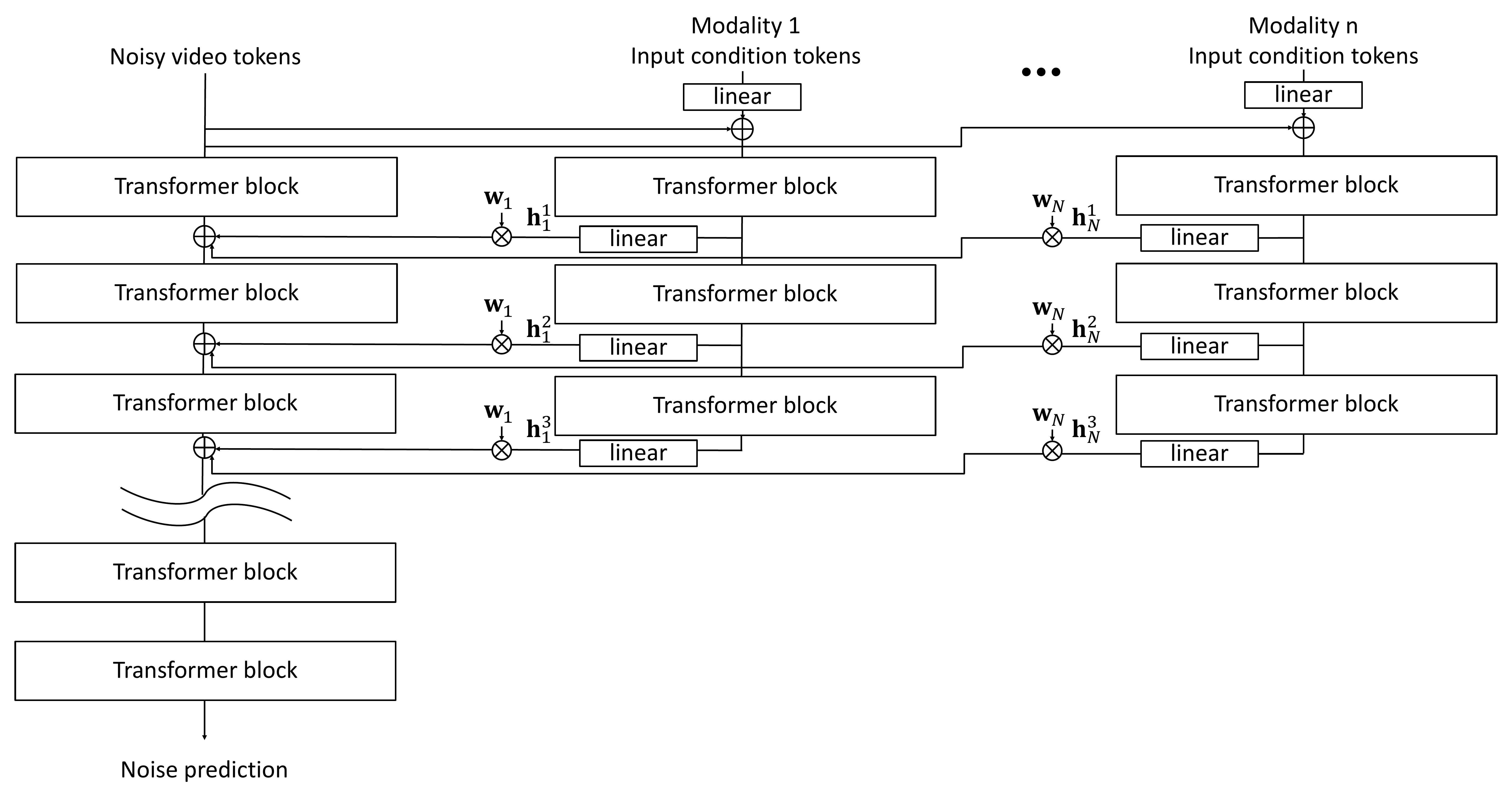}
    \caption{{\bf \modelname} is a world generator with adaptive multimodal control. It contains multiple control branches to extract control information from different modality inputs such as segmentation, depth, and edge. We apply spatiotemporal control maps $\mathbf{w} = \{\mathbf{w}_1, \mathbf{w}_2, ..., \mathbf{w}_N\}$ to weight the outputs computed by different control branches before channeling them back to the main generation branch. The spatiotemporal control map allows the model to leverage the most relevant modalities in different regions for optimal output quality.}
    \label{fig:multicontrol}
\end{figure*}

The proposed \modelname is a diffusion-based multimodal controllable world generator. It is constructed by post-training the Cosmos-Predict1 diffusion world model~\citep{nvidia2025cosmos}. Let $\mathbf{c}_1$, $\mathbf{c}_2$, ..., $\mathbf{c}_N$ be $N$ conditional inputs, representing $N$ different modalities. \modelname learns to leverage these $N$ input videos to generate the world simulation. 

For additional control, we further introduce the spatiotemporal control map $\mathbf{w}\in \mathbb{R}^{N \times X \times Y \times T}$ where $X$, $Y$, and $T$ are the video width, height, and frame counts. The spatiotemporal control map enables fine-grained, adaptive control, allowing the model to leverage the most relevant modalities in different spatiotemporal regions for optimal output quality. Let $\mathbf{h}_i^j \in \mathbb{R}^{X \times Y \times T}$ be the activations from the $j$th block of the $i$th control branch. Also, let $\mathbf{w}_i \in \mathbb{R}^{X \times Y \times T}$ be the $i$th slice of the spatiotemporal control weight. The final activation from the $j$th block of the $i$th control branch to be added back to the main branch is given by $\mathbf{w}_i \cdot \mathbf{h}_i^j$ where $\cdot$ denotes the element-wise product. We visualize our adaptive multimodal ControlNet in~\cref{fig:multicontrol}.

The spatiotemporal control maps can be derived through various approaches. First, they can be manually designed. Second, they can be adaptively derived from heuristic rules based on prior observations of the modalities. Lastly, one can train a neural module to predict the spatiotemporal control map. Note that at each spatiotemporal site, if the sum of the control maps across different modalities is greater than one, we apply normalization to the modality weights so that they sum up to one.

We train our individual control branches separately following the practice described in~\cref{sec::preliminary} and fuse them only at inference. This strategy has several advantages as compared to directly training all the control branches at once. First, it is more memory efficient, as we only need to fit one control branch into memory at a time during training. This divide-and-conquer approach helps ease the implementation burden in large-scale world model training where memory-expensive video data are used. Second, different modalities can be trained with different data, as paired data may be difficult to obtain for certain modalities. Finally, it is more flexible, as we can easily add or remove modalities at inference time after the training is done. 

\section{Modality and Training} \label{sec::training}

Our first realization of \modelname is a post-trained Cosmos-Predict1-7B-Video2World model~\citep{nvidia2025cosmos}, referred to as \sevenb. We also realize a special version of \modelname for autonomous vehicle tasks. The model is post-trained from Cosmos-Predict1-7B-Video2World-Sample-AV model~\citep{nvidia2025cosmos}, which is a finetuned version of Cosmos-Predict1-7B-Video2World on dash cam videos. We will term this model \sevenbav.

For \sevenb, we train the control branches with the high-quality finetuning dataset in~\citep{nvidia2025cosmos}. For ease of reference, when \sevenb operates on the single modality setting---the basic ControlNet setting---we will refer to it as \sevenb [modality name], such as \sevenb [Depth]. We train each control branch with 1024 NVIDIA H100 GPUs for a period of 2 to 4 weeks depending on the modality. Inherited from Cosmos-Predict1-7B-Video2World, an inference call of \sevenb generates a 5-second 1280x704p video under 24 fps, which translates to 56K tokens\footnote{We use Cosmos-Tokenize1-CV8x8x8-720p, which is a causal video tokenizer with 8x8x8 compression rate across t-axis, x-axis, and y-axis~\citep{nvidia2025cosmos}. 
$56{,}320$ tokens is computed as: $1280$ (width) $\div 8$ (tokenize) $\div 2$ (patchify) $\times 704$ (height) $\div 8$ (tokenize) $\div 2$ (patchify) $\times [(121-1) \div 8 + 1]$ (tokenized frames).} predicted by the model.

We introduce the following modalities in the paper:

\begin{figure*}[t!]
\centering
    \includegraphics[width=0.99\textwidth]{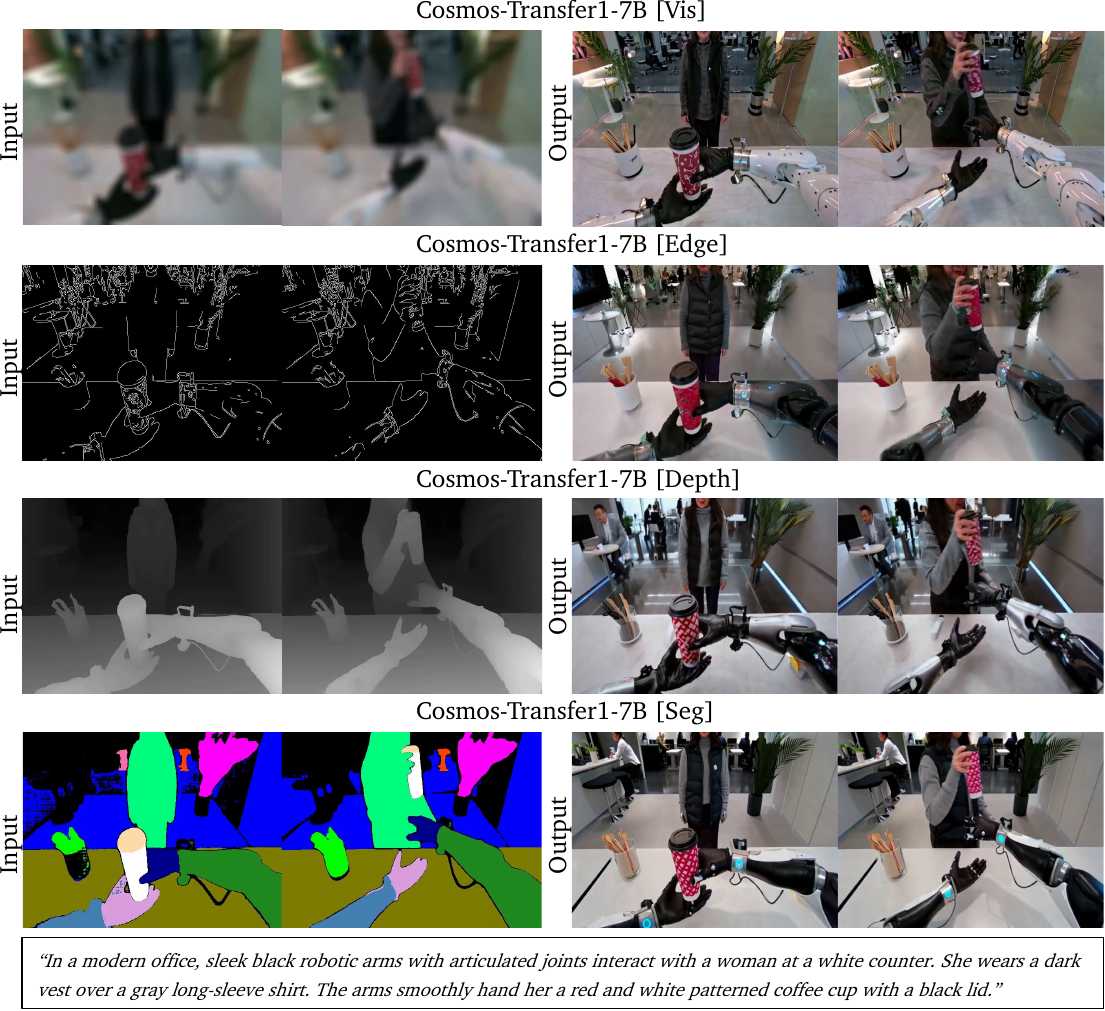}
    \caption{\textbf{Input and generated videos from \sevenb operating on individual modality settings using the same prompt.} In particular, \sevenb [Vis] preserves the colors and overall composition while altering texture details. On the other hand, \sevenb [Edge] maintains the object boundaries while changing colors. Similarly, \sevenb [Depth] preserves the scene geometry, while \sevenb [Seg] preserves the scene semantics.}
    \label{fig:ctrlnets_part1}
\end{figure*}

\begin{itemize}
    \item \textbf{Blur visual}. We apply bilateral blur~\citep{tomasi1998bilateral} to the input video to create a blurry video as the input modality. This control is useful when we plan to keep the original colors and rough shapes while only adding or changing texture details of the input. For example, it can bring realistic edge transition while improving details and clarity when we transfer CG-rendered videos to realistic videos. During training, we randomize various parameters in bilateral blur as a data augmentation strategy. We will call \sevenb operating in this setting \sevenb [Vis].
    \item \textbf{Edge}. We extract Canny edges~\citep{canny1986computational} from the input video frame-by-frame, and feed the edge video as the input modality to the control branch. This modality is useful when we only want to keep the overall structure of the scene, but give the model more freedom to creatively fill in the rest. During training, we randomize various thresholds in the Canny edge detector as a data augmentation strategy. We will call \sevenb operating in this setting \sevenb [Edge].
    \item \textbf{Depth}. We compute depth maps from the input video using DepthAnything2~\citep{depth_anything_v2} and then normalize the extracted depth to [0, 1] to generate a depth video. This control is particularly useful when keeping the 3D geometry of the input world is essential.  We will call \sevenb operating in this setting \sevenb [Depth].
    \item \textbf{Segmentation}. We use GroundingDino~\citep{liu2024grounding} to detect objects in the caption from the first frame of the video. We then use SAM2~\citep{ravi2024sam} to extract object segmentation masks from the whole video. Since there can be infinitely many object categories, we randomize the colors in the segmentation masks, \emph{i.e}\onedot colors no longer have semantic meanings and only represent different objects. This model is advantageous when we aim to keep the original segmentation layout of the scene while generating a new video with a high degree of freedom. We will call \sevenb operating in this setting \sevenb [Seg].
\end{itemize}

We use a different dataset for training \sevenbav. In autonomous driving applications, HD maps with dynamic 3D bounding boxes are commonly used perception signals, while LiDAR scans are typically collected as a complement to RGB videos. Together, these modalities provide a comprehensive semantic representation of a 3D environment in the autonomous vehicle setting. Based on this motivation, we curated a $360$-hour, high-quality autonomous driving dataset with additional HD map and 3D bounding box annotations.  We call this newly curated high-quality dataset Real Driving Scene HQ (RDS-HQ). RDS-HQ comprises of 65K 20-second surrounding-view video clips (equivalent to approximately 360 hours of videos) along with corresponding 10 Hz LiDAR scans, captured with the NVIDIA driving platform. Each clip is annotated with dense captions to enhance the alignment with domain-specific details like camera mounting position and contender vehicle density.

The modalities used by \sevenbav are listed below.

\begin{itemize} 
    \item \textbf{HDMap}. High-definition (HD) maps provide detailed and precise road annotations, including lane lines, road boundaries, stop lines, poles, crosswalks, road markings, traffic lights, and traffic signs. We constructed our HD map using a pre-built city-level LiDAR map, leveraging an in-house auto-labeling pipeline. We further augmented the HD map annotations with 3D bounding box detection and tracking of dynamic objects, including vehicles, pedestrians, and other vulnerable road users. All auto-labeled annotations undergo rigorous human quality assessment (QA) to ensure high accuracy. The HDMap ControlNet is particularly advantageous when preserving the original road layout of a driving scene while generating a new video with a high degree of freedom. Additionally, it plays a crucial role in simulating different ego-car trajectories, as HD maps provide precise information on the relative pose translation between the ego camera and the road geometry. We will call \sevenbav operates in this setting \sevenbav [HDMap].
    
    \item \textbf{LiDAR}. To synchronize the collected 10 FPS LiDAR scans with the 30 FPS camera frames, we temporally interpolate and densify the LiDAR data. For each camera frame, we select the nearest LiDAR scan along with four additional adjacent scans (two preceding and two subsequent) to provide enhanced temporal context. Static points from these LiDAR scans are directly projected onto the camera frame, while for the dynamic points, identified using the object detection auto-labeling pipeline, we adjust their positions by interpolating the 3D bounding boxes prior to projection. To further reduce projection-induced holes, we apply a kernel-based interpolation with a kernel size of 4.  The LiDAR ControlNet can be utilized to preserve the original semantic details of the driving scene while allowing modifications to environmental conditions and lighting. We will call \sevenbav operates in this setting \sevenbav [LiDAR].
  
\end{itemize}

\begin{figure*}[h!]
\centering
    \includegraphics[width=0.99\textwidth]{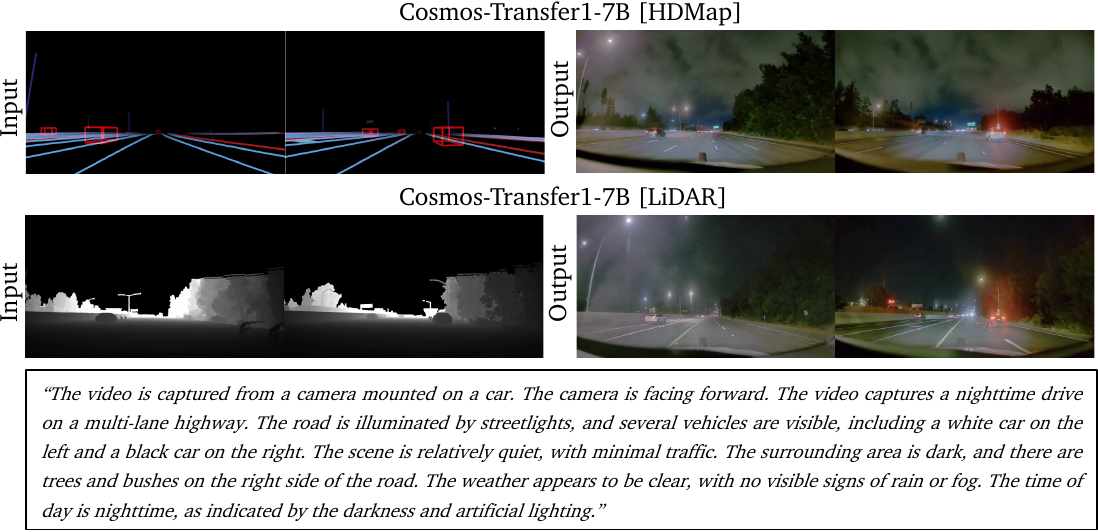}
    \caption{\textbf{Input and generated videos from \sevenbav operating on individual modality settings.} \sevenbav [HDMap] preserves the original road layout of a driving scene while \sevenbav [LiDAR] preserves the input semantic details.}
    \label{fig:ctrlnets_part2}
\end{figure*}

\begin{figure*}[t!]
\centering
    \includegraphics[width=0.99\textwidth]{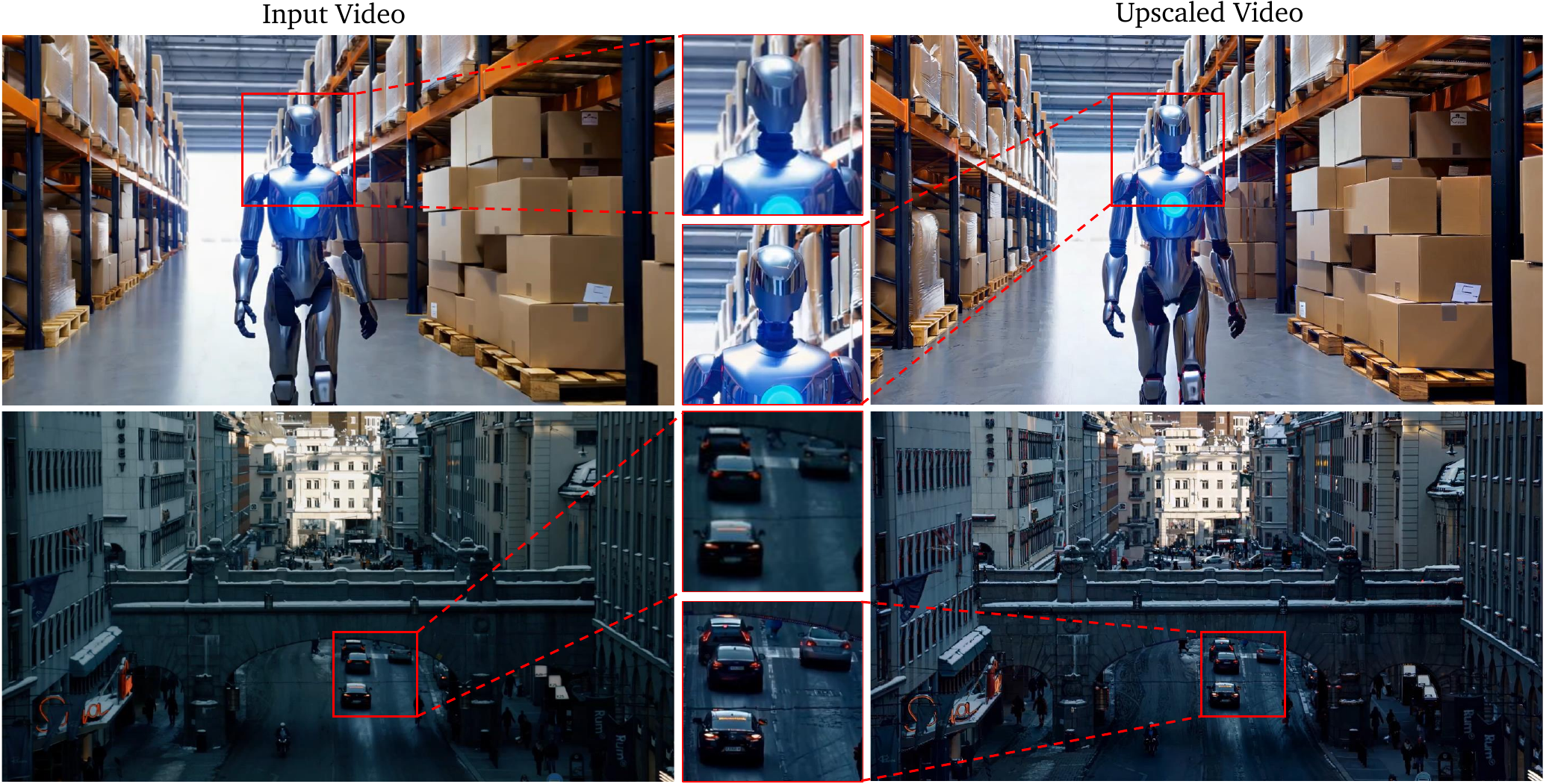}
    \caption{\textbf{\sevenbupscaler upscales videos from 720p to 4k resolution.} The input video in the first row is a generated video, while the second row is a real video. Note how the model adds realistic reflections and sharpens the textures in the input.}
    \label{fig:ctrlnets_part3}
\end{figure*}

Finally, we also train an Upscale ControlNet to upscale the generated videos from 720p to 4k resolution. We show examples of the 4KUpscaler in~\cref{fig:ctrlnets_part3}. We take random crops of corrupted high-resolution videos using the corruption techniques in Real-ESRGAN~\citep{wang2021real} as input. We then train the ControlNet to recover the original high-quality video patches. During inference, we adopt patch-based generation and divide the 4k output into $3\times 3$ grids with overlapping regions. At each denoising step, we run inference for each grid and average their results in the overlapping regions. This ensures the output video is seamless around the boundaries. We will call the 4KUpscaler version \sevenbupscaler.

\section{Evaluations} \label{sec::evaluation}
In this section, we present a comprehensive evaluation of \modelname. We begin by introducing the dataset and evaluation metrics. In Sec.~\ref{sec::evaluation::uni_vs_multi}, we first show basic characteristics of the models by comparing single control models with multimodal control models that use uniform spatial weights. 
In Sec.~\ref{sec::st_weight_design_example}, we present an approach to automatically create spatiotemporal control maps, and analyze the characteristics of \modelname when utilizing them. 
Finally, Sections \ref{sec:eval:robotics} and \ref{sec:eval:av} provide case studies demonstrating the applicability of our method for Sim2Real data generation in robotics and autonomous driving, respectively.

{\bf \evalsetname.} To evaluate the characteristics of \modelname, we curate \textit{\evalsetname} --- an evaluation dataset consisting of 600 examples across three key scenarios: robotic arm operations, driving, and ego-centric everyday life scenes, each representing a critical aspect of Physical AI. Specifically, we randomly sample 200 examples from the \textbf{AgiBot World} dataset~\citep{contributors2025agibotworld} for robotic arm operations, which capture fine-grained manipulation, dexterity, and interaction with objects, essential for automation and industrial robotics. For driving scenarios, we sample 200 examples from the \textbf{OpenDV} dataset~\citep{yang2024genad}, representing mobile autonomy, perception in dynamic environments, and decision-making in complex traffic conditions. Lastly, for ego-centric everyday life scenes, we sample 200 examples from the \textbf{Ego-Exo-4D} dataset~\citep{grauman2024egoexo4dunderstandingskilledhuman}, focusing on human-centric AI, interaction modeling, and embodied perception in unstructured settings. This selection provides a diverse evaluation suite, covering both structured and unstructured real-world environments that are handy for benchmarking \modelname performance.

{\bf Metrics.} We evaluate \modelname based on three key aspects: 1) Adherence to Control Input Signals, 2) Generation Diversity, and 3) Overall Generation Quality.

\begin{itemize}  
    \item \textbf{Adherence to Control Input Signals}. For quantitative evaluation, we first transform both the input sample videos and the generated videos into a shared representation space by applying operations, including blurring, edge extraction, depth estimation, and semantic segmentation. We then compare these transformed representations to measure alignment, assessing whether the generated worlds accurately reflect the intended conditioning. Specifically, for the  alignment with each of the control signals, we perform the following:
    \begin{itemize}
        \item \textbf{Vis Alignment}. We apply the same blurring operation to both the input sample videos and the generated videos, and compute their Structural Similarity Index Measure (SSIM)~\citep{ssim}. We then average the scores over all the samples within the dataset. We call this metric \textbf{Blur SSIM}. Higher values mean better alignment.
        
        \item \textbf{Edge Alignment}. We apply the same Canny edge extraction to both the input sample videos and the generated videos, and compute a classification F1 score~\citep{vanRijsbergen1979} on the black-and-white pixel classification. We then average the F1 scores over the dataset. We call this metric \textbf{Edge F1}. Higher values mean better alignment.
        
        \item \textbf{Depth Alignment}. We compute the scale-invariant Root Mean Squared Error (si-RMSE)~\citep{eigen2014depth} between the depth maps extracted from both the input sample videos and the generated videos using DepthAnythingV2~\citep{depth_anything_v2}. We call this metric \textbf{Depth si-RMSE}. Lower values mean better alignment.
        
        \item \textbf{Segmentation Alignment}. We compute the mean Intersection over Union (mIoU) between the segmentation masks generated using our GroundingDINO+SAM2 annotation pipeline~\citep{ravi2024sam}. Since GroundingDINO is an open-set object detection model, it might detect the same object multiple times, resulting in repeated instance segmentations. To address this, we aggregate all segmentation masks associated with the same phrase in the captions into a single segmentation mask, ensuring comparisons are made at the object level across the video. Next, we establish correspondences between the ground truth and generated segmentation masks using an matching algorithm based on the IoU distance function. Masks with an IoU below 0.1 are discarded to eliminate spurious matches, and the final mIoU score is computed as the mean IoU over all valid correspondences. We call this metric \textbf{Mask mIoU}. Higher values mean better alignment.
        
    \end{itemize}

    \item \textbf{Diversity}. For a given condition input, we expect \modelname can generate different videos based on different text prompts. To evaluate diversity, we design the following metric based on LPIPS~\citep{lpips}. For each condition input, we generate $K$ videos from $K$ different text prompts. We compute the LPIPS scores for the $K(K-1)/2$ pairs of the $K$ videos. We average the LPIPS scores over all these pairs to get the averaged perceptual similarity. We then average over all samples in the dataset. We call this metric \textbf{Diversity-LPIPS}. Higher values mean better diversity.
    
    \item \textbf{Overall Generation Quality}. We report the DOVER-technical score~\citep{wu2023exploring}, which serves as a perceptual metric for assessing the general aesthetic quality of the generated videos. A higher DOVER score indicates better visual quality. We average the DOVER scores across all videos within the dataset. We call this metric \textbf{Quality Score}.
\end{itemize}

\subsection{Unimodal versus Multimodal}
\label{sec::evaluation::uni_vs_multi}

Before introducing the spatiotemporal control map, we first characterize the impact of different control modalities by evaluating several special cases of \modelname: 
\begin{itemize}
    \item \textit{Single control model.} In the single-control setting, the model is conditioned on only one control modality: blurred original video, Canny edges, depth maps, or segmentation maps, while the others are disabled. These settings help understand the strength of individual modalities.
    
    \item \textit{Multimodal control model with one single modality excluded.} In this setting, except one modality, all other modalities contribute equally. These settings help understand the weakness of the overall model when one modality is excluded.
    
    \item \textit{Full multimodal control model with uniform weights.} In this setting, we fuse all the modalities with an equal weight across the spatiotemporal control map.
\end{itemize}

\cref{tab:eval_general_uniform} shows the experiment results. As expected, the vis control model achieves the highest Blur SSIM (0.96), reflecting its strength in preserving coarse structure and color.
Similarly, the edge control model attains the best Edge F1 score (0.28), underscoring its ability to capture dense structural details in the control input\footnote{The Edge F1 score is a very strict metric that requires pixel-level alignment of the extracted thin canny edges. This strictness contributes to relatively low F1 scores for all methods.}. 
The depth control and segmentation control models do not achieve the highest scores in their respective metrics; the depth control model records a Depth si-RMSE of 0.49 (comparable to the blur visual control model), while the segmentation control model achieves a Mask mIoU of 0.68. As shown in \cref{fig:ctrlnets_part1} and~\cref{fig:spatiotemporal-diagram}, depth maps can be largely homogeneous in distant background regions, while segmentation maps from SAM2 often comprise broad color regions rather than fine-grained structures. This sparsity imposes fewer constraints on the generation process, giving the model additional freedom that may lead to less accurate reconstructions when these modalities are used in isolation. 

Similar patterns can be observed when we only exclude a single modality and assign uniform weights to other modalities. For example, when only blur visual control is excluded  (\textit{\sevenb Uniform Weights, no Vis}), it achieves the lowest Blur SSIM (0.68) compared to others, indicating that the dense structural cues provided by the blur visual input are critical for high alignment. Notably, the diversity scores reveal that modalities with denser structural information (Blur visual and Edge) tend to yield lower diversity, whereas those with sparser cues (Depth and Segmentation) promote higher diversity. Specifically, excluding Blur visual or Edge increases the diversity LPIPS to 0.37 and 0.31, respectively, while removing Depth or Segmentation reduces it to 0.25 and 0.23, respectively. 
The above observations suggest that the Blur visual and Edge controls, by providing dense structural information, are particularly useful for tasks requiring precise alignment and high-fidelity output---such as photorealistic style transfer or Sim2Real transformation in controlled environments. In contrast, Depth and Segmentation controls, which offer sparser structural guidance, allow for greater output variability and are thus suited for applications where diversity is desired, such as novel scene synthesis for data augmentation.

In contrast, the adaptive multimodal control model (\textit{\sevenb Uniform Weights}), which leverages multiple control inputs, consistently produces high-quality results. Although it ranks second in both Vis and Edge Alignment---likely because each modality is assigned a lower weight (0.25) compared to 1.0 in the single-control setting, it achieves the best depth reconstruction and the highest Quality Score (8.54). This suggests that, while individual control inputs excel at capturing specific aspects of the video, our multimodal control strategy effectively integrates complementary features from all modalities, leading to a more balanced and accurate output. These results validate our motivation for the multimodal control design of \modelname. In the following sections, we introduce a method for automatically generating spatially-varying control weights and analyze how \modelname behaves when applying region-specific control.

\begin{table}[t!]
\centering
\caption{\textbf{Quantitative evaluation on \evalsetname for various \modelname configurations.} We compare single control models (each conditioned on a single modality) with multimodal variants that use spatially uniform weights. For the multimodal cases, ``\sevenb, Uniform Weights'' denotes the full model that integrates all four control modalities (each weighted at 0.25), while variants such as ``\sevenb, Uniform Weights, No Vis'' exclude a specific modality (here, the blur visual control), with the remaining modalities retaining equal weights. Best results are in bold; second-best are underlined.}
\label{tab:eval_general_uniform}
\resizebox{\textwidth}{!}{
\begin{tabular}{l|cccc|cc}
\toprule
\multicolumn{1}{c|}{\multirow{4}{*}{Model}} & \makecell{Vis \\Alignment} & \makecell{Edge\\Alignment} & \makecell{Depth\\Alignment} & \makecell{Segmentation\\Alignment} & Diversity & \makecell{Overall\\Quality} \\
\cmidrule(lr){2-7}
          & \makecell{Blur\\SSIM $\uparrow$} & \makecell{Edge\\F1 $\uparrow$} & \makecell{Depth\\si-RMSE $\downarrow$} & \makecell{Mask\\mIoU $\uparrow$} & \makecell{Diversity\\LPIPS $\uparrow$} & \makecell{Quality\\Score $\uparrow$}\\
\midrule
\sevenb [Vis] & \textbf{0.96} & 0.16 & 0.49 & \textbf{0.72} & 0.19 &  5.94 \\
\sevenb [Edge] & 0.77 & \textbf{0.28} & 0.53 & 0.71 & 0.28 & 5.48  \\
\sevenb [Depth] & 0.71 & 0.14 & 0.49 & 0.70 & \underline{0.39} & 6.51 \\
\sevenb [Seg] & 0.66 & 0.11 & 0.75 & 0.68 & \textbf{0.42} & 6.30 \\
\sevenb Uniform Weights, no Vis & 0.68 & 0.13 & 0.57 & 0.67 & 0.37 & \underline{8.02}  \\
\sevenb Uniform Weights, no Edge & 0.81 & 0.10 & 0.53 & 0.66 & 0.31 & 7.68  \\
\sevenb Uniform Weights, no Depth & 0.83 & 0.15 & 0.52 & 0.69 & 0.25 & 7.49  \\
\sevenb Uniform Weights, no Seg & 0.84 & 0.15 & \textbf{0.43} & 0.70 & 0.23 & 7.83 \\
\sevenb Uniform Weights & \underline{0.87} & \underline{0.20} & \underline{0.47} & \textbf{0.72} & 0.22 & \textbf{8.54} \\
\bottomrule
\end{tabular}
}
\end{table}

\subsection{Case Study for Spatiotemporal Control Maps}\label{sec::st_weight_design_example}

\begin{figure}[t!]
    \centering
    \includegraphics[width=\textwidth]{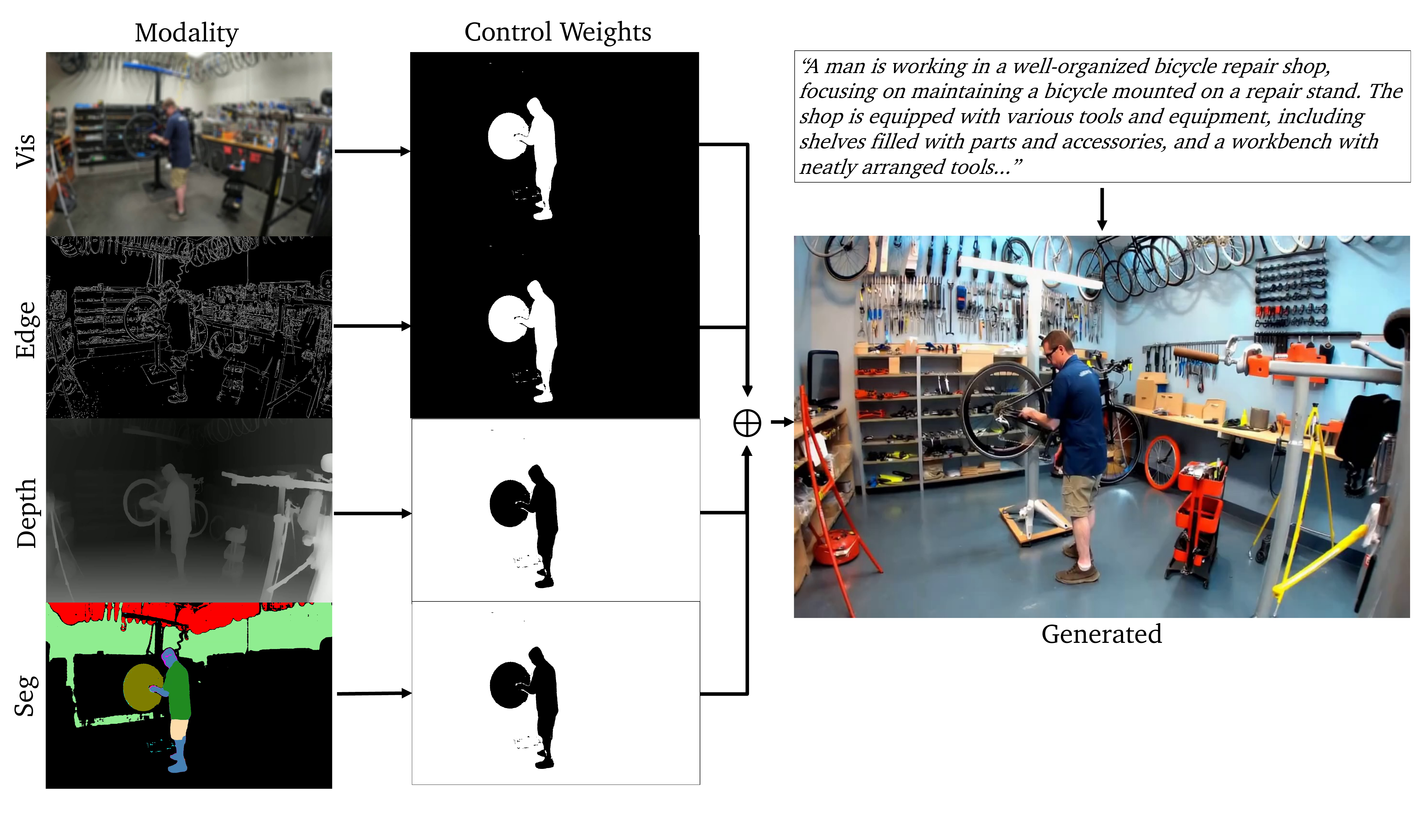}
    \caption{\textbf{Diagram of spatiotemporal control weighting by different modalities} (Vis, Edge, Depth and Segmentation). The control weight maps are $0.0$ in black pixel areas, and $0.5$ in white areas. We note that while the caption broadly specifies a bicycle repair shop scene, the blue shirt with a white logo and the skin color of the man are maintained, due to these pixels being controlled by Vis and Edge. On the other hand, for the background controlled by Depth and Segmentation, the objects are positioned in the scene consistently but have their colors and textures randomized (e.g. red toolbox, yellow tripod, white repair stand). A new tool rack on the wall on the right is also added by the model.
}
    \label{fig:spatiotemporal-diagram}
\end{figure}

To demonstrate the strength of the proposed spatiotemporal control map, we design a \textit{SalientObject} algorithm where we give different modalities different weights based on whether the location is from foreground or background. Specifically, we prompt a VLM~\citep{openai2024gpt4technicalreport} to classify each pre-extracted GroundingDINO+SAM2 mask into either foreground and background. To do so, we provide the model with (1) reference video frames, (2) the complete caption we use for generation, and (3) the list of phrases associated to each segment mask, and then prompt the model to classify whether each phrase belongs to a salient foreground object in the scene.

Following this procedure, we can create a spatiotemporal foreground-background control weight mask. We show an example in \cref{fig:spatiotemporal-diagram}, where we assign a weight of $0.5$ to vis and edge modalities respectively in the foreground, and a weight of $0.5$ to depth and segmentation respectively in the background. While a multitude of heuristics could be used, we choose this composition strategy to exemplify a situation in which the user wants to have close resemblance to the original videos for the salient objects in the scene but generate more diverse videos in the background. Since vis and edge modalities are more constraining than depth and segmentation modalities, one can achieve this by using vis and edge modalities for the foreground, which leads to lower degree-of-freedom, and depth and segmentation modalities for the background, which have higher degree-of-freedom.

\begin{figure}[t!]
    \centering
    \includegraphics[width=\textwidth]{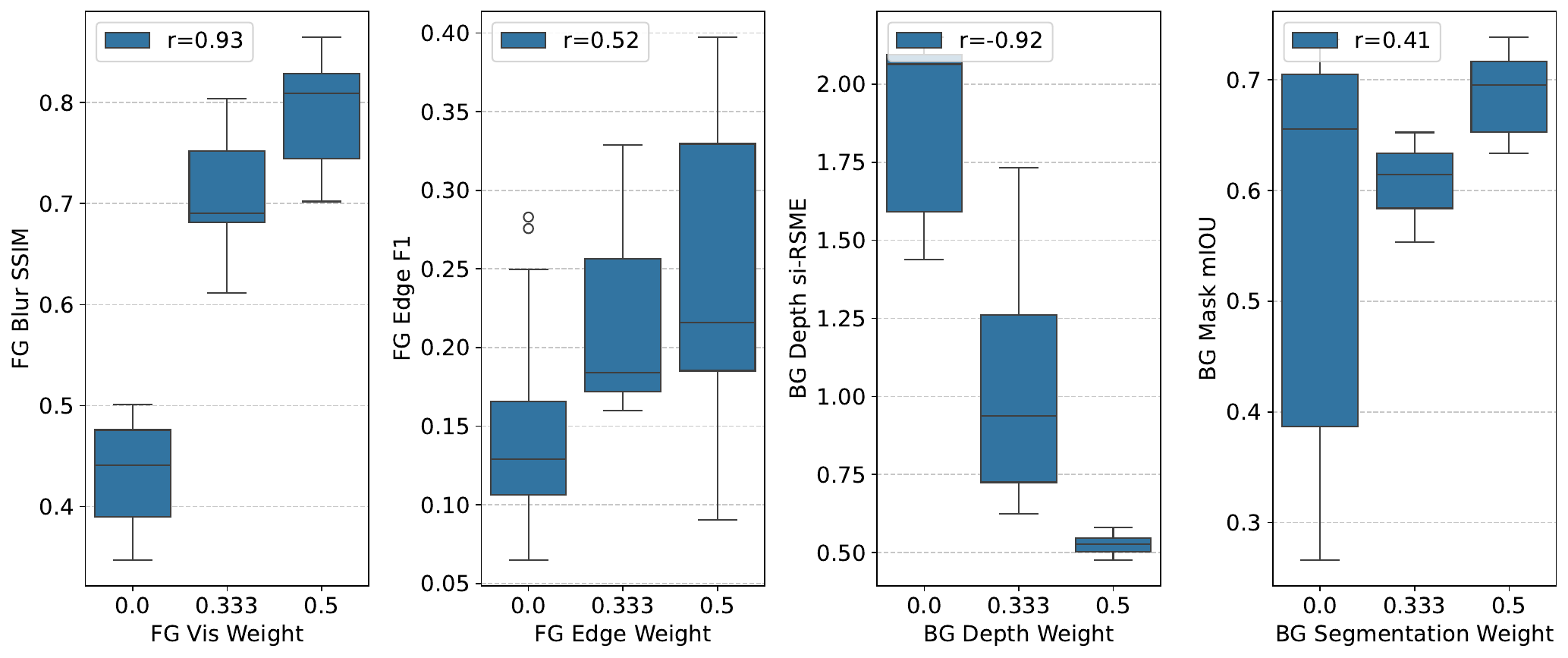}  
    \caption{Correlations of modality weights on foreground (FG) region (for Vis and Edge) or background (BG) region (for Depth and Segmentation) with ground truth modality.}
    \label{fig:overall-spatiotemporal}
\end{figure}

In addition to this qualitative example, we also conduct a series of quantitative experiments. The first experiment to demonstrate granular spatiotemporal control is shown in \cref{fig:overall-spatiotemporal}. Here, we condition the foreground with edge and vis, and background with depth and segmentation. We ablate the conditioning strength per modality ($0$, $0.333$, $0.5$) and show a strong correlation to the relevant alignment metric in the affected pixel regions. For example, by selectively increasing the foreground vis weight, we improve the foreground blur SSIM from $0.43$ to $0.81$, with a Pearson correlation coefficient of $0.93$. Similarly, by selectively increasing the background depth weight, we improve the Depth si-RSME from $1.88$ to $0.52$, with a Pearson correlation coefficient of $-0.92$.

\begin{table}[t]
\centering
\caption{\textbf{Quantitative evaluation on \evalsetname for \modelname-7B with spatiotemporal weights derived from our \textit{SalientObject} algorithm.} The leftmost eight columns specify the weight design for the four modalities respectively. For each metric, ``FG'' denotes the result in that metric computed in the foreground region, and ``BG'' stands for background.}
\label{tab:eval_general_adaptive}
\resizebox{\textwidth}{!}{
    \begin{tabular}{cccc|cccc|cccccccc|ccc}
    \toprule
    \multicolumn{8}{c|}{Control Weights} & \multicolumn{2}{c}{\makecell{Vis\\Alignment}} & \multicolumn{2}{c}{\makecell{Edge\\Alignment}} & \multicolumn{2}{c}{\makecell{Depth\\Alignment}} & \multicolumn{2}{c|}{\makecell{Segmentation\\Alignment}} & \multicolumn{2}{c}{Diversity} & \makecell{Overall\\Quality} \\
    \midrule
    \multicolumn{4}{c|}{FG} & \multicolumn{4}{c|}{BG} & \multicolumn{2}{c}{\makecell{Blur\\SSIM $\uparrow$}} & \multicolumn{2}{c}{\makecell{Edge\\F1  $\uparrow$}} & \multicolumn{2}{c}{\makecell{Depth\\si-RSME $\downarrow$}} & \multicolumn{2}{c|}{\makecell{Mask\\mIoU $\uparrow$}} & \multicolumn{2}{c}{\makecell{Diversity\\LPIPS $\uparrow$}} & \makecell{Quality\\Score $\uparrow$} \\
    Vis & Edge & Depth & Seg & Vis & Edge & Depth & Seg & FG & BG & FG & BG & FG & BG & FG & BG & FG & BG & \\
    \midrule
    0.5 & 0.5 & 0& 0 & 0 & 0 & 0.5 & 0.5 & \textbf{0.81} & 0.71 & \textbf{0.27} & 0.14 & \textbf{0.37} & 0.52 & \underline{0.77} & 0.68 & 0.01 & \textbf{0.33} & \textbf{8.29} \\
    0 & 0 & 0.5 & 0.5 & 0.5 & 0.5 & 0 & 0 & 0.68 & \textbf{0.93} & 0.17 & \textbf{0.25} & 0.38 & \textbf{0.40} & \underline{0.77} & \textbf{0.75} & \textbf{0.12} & 0.03 & 8.08 \\
    \bottomrule
    \end{tabular}
}
\end{table}

The second experiment we perform is shown in \cref{tab:eval_general_adaptive}.  Here, we use two different strategies to determine the spatiotemporal control map: 
\begin{itemize}
    \item We condition the foreground with edge and vis, and background with depth and segmentation.
    \item We condition the foreground with depth and segmentation, and background with edge and vis, essentially inverting foreground/background.
\end{itemize}

From \cref{tab:eval_general_adaptive}, we observe that when swapping the edge and vis conditioning from foreground to background, there is a marked improvement in Blur SSIM and Edge F1 in the background and a degradation in the foreground. We also observe that vis and edge conditioning perform competitively for depth and segmentation alignment, in agreement with \cref{tab:eval_general_uniform}.

Furthermore, we observe in \cref{tab:eval_general_adaptive} that when we swap the (foreground or background) region from vis and edge conditioning (low degree-of-freedom) to depth and segmentation conditioning (high degree-of-freedom), we observe a marked increase in diversity LPIPS score (FG $0.01\rightarrow0.12$, BG $0.03\rightarrow0.33$) and on-par visual quality, while also maintaining good depth and segmentation alignment (especially in reference to unconditioned baselines in \cref{fig:overall-spatiotemporal}). This matches our expectation from the qualitative example shown in \cref{fig:spatiotemporal-diagram}.

\subsection{Case Study for Robotics Sim2Real Data Generation}\label{sec:eval:robotics}

In robotics research, the availability of large-scale, high-quality data is critical, as supported by established scaling laws \citep{lin2025datascalinglawsimitation}. While simulation provides a means to generate extensive datasets, the substantial domain gap between synthetic and real-world data presents significant challenges in transferring models trained in simulated environments to real-world applications \citep{muratore2022robot}. 

To evaluate \modelname for robotics data generation, we curated a small simulated dataset of 20 robot manipulation scenarios in a basic kitchen scene using NVIDIA Omniverse and Isaac Lab. The kitchen scenes were programmatically generated, and contain furniture with articulations for the robot to interact with. The tasks for the robot include opening and closing cabinets, and picking and placing common kitchen objects. The robot motions were computed by an automated task and motion planning algorithm~\citep{garrett2020pddlstream}. For each of the scenarios, we prepared six different text prompts with descriptions of different illumination, scene details, and operating environments. In addition to RGB videos, we also output segmentation and depth maps for each scenario. In addition to the aforementioned four single-modal \sevenb models, in order to fully preserve robot dynamics, we also employed the \sevenb model with the spatiotemporal control map. Specifically, we first extracted the foreground robot masks using semantic segmentation results and then designed different modal constraint settings for both the foreground and background. 

\begin{table}[h]
    \centering
    \footnotesize
    \caption{\textbf{Quantitative evaluation of \modelname on robotics Sim2Real data generation task}, including single-control models and two multimodal control variants with different spatiotemporal control maps. Best results are in bold; second-best are underlined.}
    \resizebox{\textwidth}{!}{
    \begin{tabular}{l|ccccc|cc}
        \toprule
        \multicolumn{1}{c|}{\multirow{4}{*}{Model}} & \makecell{Vis \\Alignment} & \makecell{Edge\\Alignment} & \makecell{Depth\\Alignment} & \makecell{Segmentation\\Alignment} & \makecell{FG Segmentation\\Alignment} & Diversity & \makecell{Overall\\Quality} \\
        \cmidrule(lr){2-8}
          & \makecell{Blur\\SSIM $\uparrow$} & \makecell{Edge\\F1 $\uparrow$} & \makecell{Depth\\si-RMSE $\downarrow$} & \makecell{Mask\\mIoU $\uparrow$} & \makecell{FG Mask\\mIoU $\uparrow$} & \makecell{Diversity\\LPIPS $\uparrow$} & \makecell{Quality\\Score $\uparrow$}\\
        \midrule
        \sevenb [Vis]  &  \textbf{0.95} & \underline{0.19} &  \textbf{0.82}  &  \textbf{0.65} & 0.56 & 0.20 & 9.11  \\
        \sevenb [Edge]     &  0.63 & \textbf{0.40} &  1.01  &  \underline{0.63} & 0.57 & 0.36 & 7.70  \\
        \sevenb [Depth]      &  \underline{0.66} & 0.13 & \underline{0.84}  & 0.59 & 0.57  & 0.43 & 9.17 \\
        \sevenb [Seg]  & 0.47 & 0.10 & 1.34  & 0.55 & 0.54 & \textbf{0.60} & 9.29 \\
        \sevenb, Setting1   & 0.51 & 0.12 & 1.30 & 0.59 & \underline{0.61}  & 0.57 & \underline{9.57} \\
        \sevenb, Setting2    & 0.50 & 0.14  & 1.41 & 0.60 & \textbf{0.63} & \underline{0.58} & \textbf{10.42} \\
        \bottomrule
    \end{tabular}
    }
    \label{tab:eval_robot}
\end{table}

\begin{figure}[h!]
    \centering
    \includegraphics[width=\textwidth]{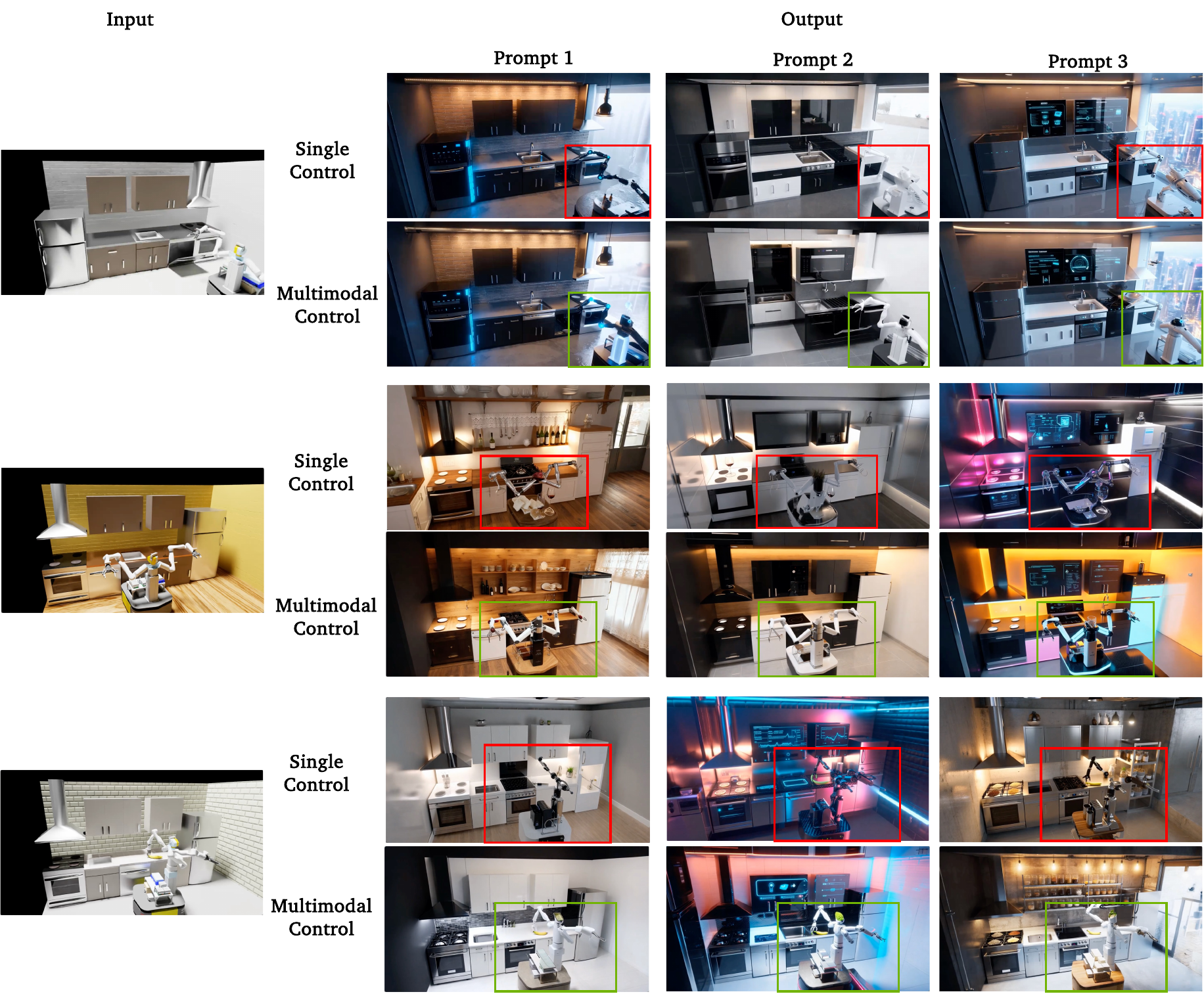}  
    \caption{\textbf{Example results of \modelname for robotic data generation.} The left column displays input videos generated by NVIDIA Isaac Lab, while the right three columns show results from \sevenb with different condition modalities and spatiotemporal control maps. For each example, the top row (single) uses Segmentation as the condition modality with an overall constraint weight of 1. The bottom row combines Segmentation, Edge, and Vis as conditions, applying a spatiotemporal control map scheme. Specifically, a combination of Edge, Segmentation and Vis are used with a customized control weight on the foreground (robot region), while only segmentation with a control weight of 1 is applied to the background. These results demonstrate that \sevenb with the spatiotemporal control map enhances the fidelity of the foreground robot.}
    \label{fig:eval_robot}
\end{figure}

We propose two straightforward configurations for using \sevenb. In Setting 1, our goal is to preserve both the shape and appearance of the foreground robot while modifying the background. To achieve this, we apply the edge and vis controls to the foreground (FG) while using the Segmentation control for the background (BG), assigning the following weights: $w_{\text{Edge}}(FG) = 1$, $w_{\text{Vis}}(FG) = 1$, and $w_{\text{Seg}}(BG) = 1$, with all other weights set to zero. In Setting 2, we aim to maintain only the shape of the robot while allowing variations in its appearance (e.g., color and texture). Therefore, we apply only the edge control to the foreground, while still using the segmentation control for the background, with the weights:
$w_{\text{Edge}}(FG) = 1$, and $w_{\text{Seg}}(BG) = 1$, and all other weights set to zero.

\begin{figure}[t]
    \centering
    \includegraphics[width=\textwidth]{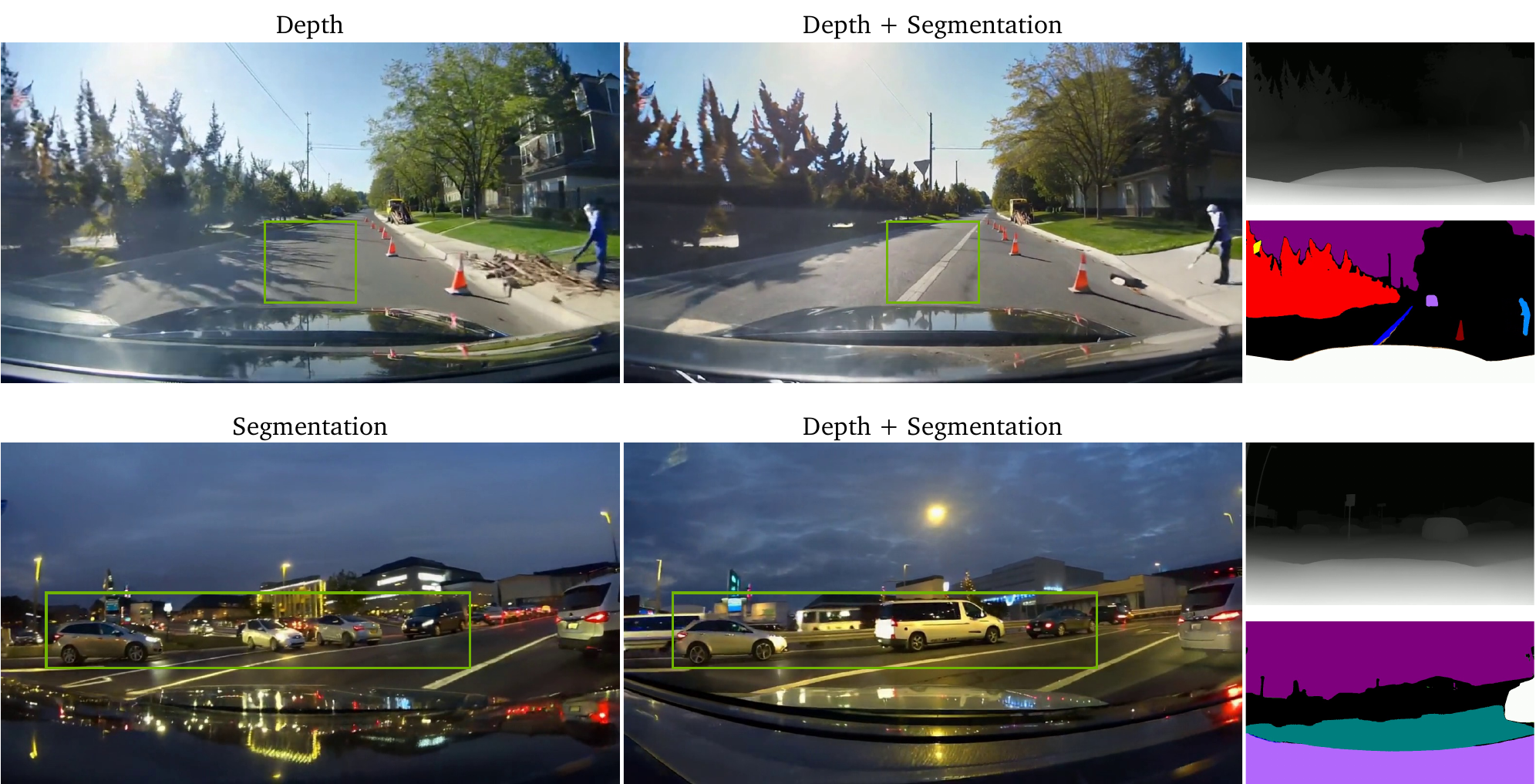}  
    \caption{\textbf{Comparison of the generation results conditioned on depth and segmentation of \sevenb}. In each example, the highlighted regions illustrate the enhancements achieved by incorporating multiple control signals over relying on a single one.}
    \label{fig:av_depth_segment}
\end{figure}
\begin{figure}[t!]
    \centering
    \includegraphics[width=1\textwidth]{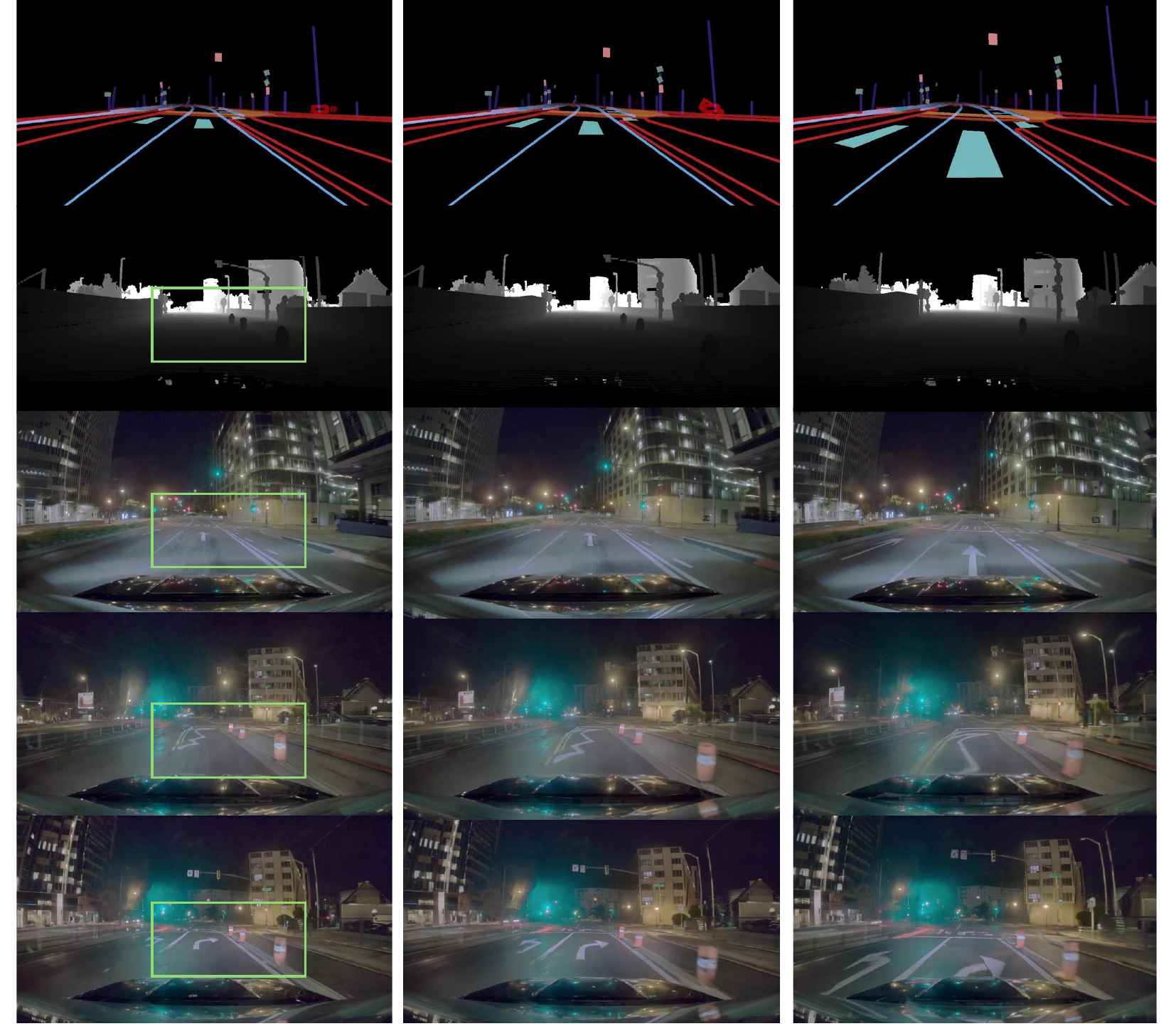}
    \caption{\textbf{Comparison of the generation results conditioned on HDMap and LiDAR of \sevenbav.} The highlighted regions illustrate the enhancements achieved by incorporating multiple control signals compared to relying on a single one. \textbf{1st row}: HDMap condition. \textbf{2nd row}: LiDAR condition. \textbf{3rd row}: Video generated using only HDMap. \textbf{4th row}: Video generated using only LiDAR, where traffic cones are introduced by LiDAR, but lane markings are incorrect. \textbf{5th row}: Video generated using both HDMap and LiDAR, where the lane layout is improved and more detailed objects are synthesized.}
    \label{fig:driving_compare}
\end{figure}

Table~\ref{tab:eval_robot} presents the quantitative evaluation of \modelname on robotics simulation data (120 videos, i.e., $20 \times 6$). The evaluation metrics include Blur SSIM, Edge F1, Mask mIoU, Diversity-LPIPS, and Quality Score. Additionally, we compute FG Mask mIoU for the foreground robot, excluding the background, to evaluate the consistency between simulated and generated videos. As observed, single-modal \sevenb models tend to yield higher scores for the corresponding metrics (e.g., \sevenb [Vis] achieves the highest Blur SSIM) but result in a lower Quality Score than those from \sevenb under the adaptive multimodal control settings. \sevenb[Seg] produces higher Quality Scores and the highest Diversity-LPIPS but exhibits lower performance in Blur SSIM, Edge F1, and Mask mIoU, which suggests that \sevenb[Seg] can generate diverse backgrounds but may introduce artifacts into the foreground robot. Overall, the two \sevenb model settings are among the top three in Quality Score, Diversity-LPIPS, and FG Mask mIoU, while offering a more balanced performance across Blur SSIM, Edge F1, and Mask mIoU, which shows \sevenb with spatiotemporal control map has better overall video quality, diversity, and the preservation of foreground robots. These statistical findings are further illustrated in the following figure.

Figure~\ref{fig:eval_robot} shows three examples of using \sevenb to augment simulated robotics data. The left column displays the input videos generated by NVIDIA Omniverse and Isaac Lab, while the right three columns show the outputs of \sevenb with different condition modalities, spatiotemporal control maps, and text prompts. First, these results show \sevenb significantly improves the photorealism and diversity of the robotics videos, by adding more scene details and complex shading and natural illumination. Moreover, as highlighted in the red and green rectangle regions, compared to using single-modal models, we found using the full \sevenb model with spatiotemporal control maps on edge, segmentation and vis can better preserve the robot shape, reduce broken artifacts and result in overall better video quality. Please refer to the figure caption for more details.

As shown, \sevenb offers a promising approach to mitigating the domain gap between simulation and real-world by enhancing the realism and diversity of synthetic data while preserving task-relevant properties. For instance, in robotics manipulation scenarios, \sevenb can refine simulated videos by adjusting lighting, color, texture, and fine-grained scene details while maintaining physically plausible robot dynamics. Additionally, \sevenb can enrich scene complexity by introducing novel background objects, thereby increasing the ecological validity of the data. By leveraging \sevenb, researchers can improve the robustness and generalization of models trained in simulation, facilitating more effective deployment in real-world robotic environments.

\subsection{Case Study for Autonomous Driving Data Enrichment}\label{sec:eval:av}

Unlike robotics, the field of autonomous driving continuously accumulates vast amounts of real-world data collected by mass-produced vehicles and self-driving fleets. However, this data follows a highly long-tailed distribution, thus making it crucial to amplify the utilization of safety-critical corner cases. By leveraging \modelname, autonomous vehicle developers can maximize the utility of real-world edge cases, ultimately leading to enriched and diversified driving data. Specifically, one can condition \modelname on the interventions described in scenario formats to generate numerous useful visual variations for autonomous vehicle (AV) testing and training. This capability allows systematic exploration of challenging and rare scenarios, enhancing both testing coverage and training effectiveness.

\begin{figure}[t!]
    \centering
    \includegraphics[width=1\textwidth]{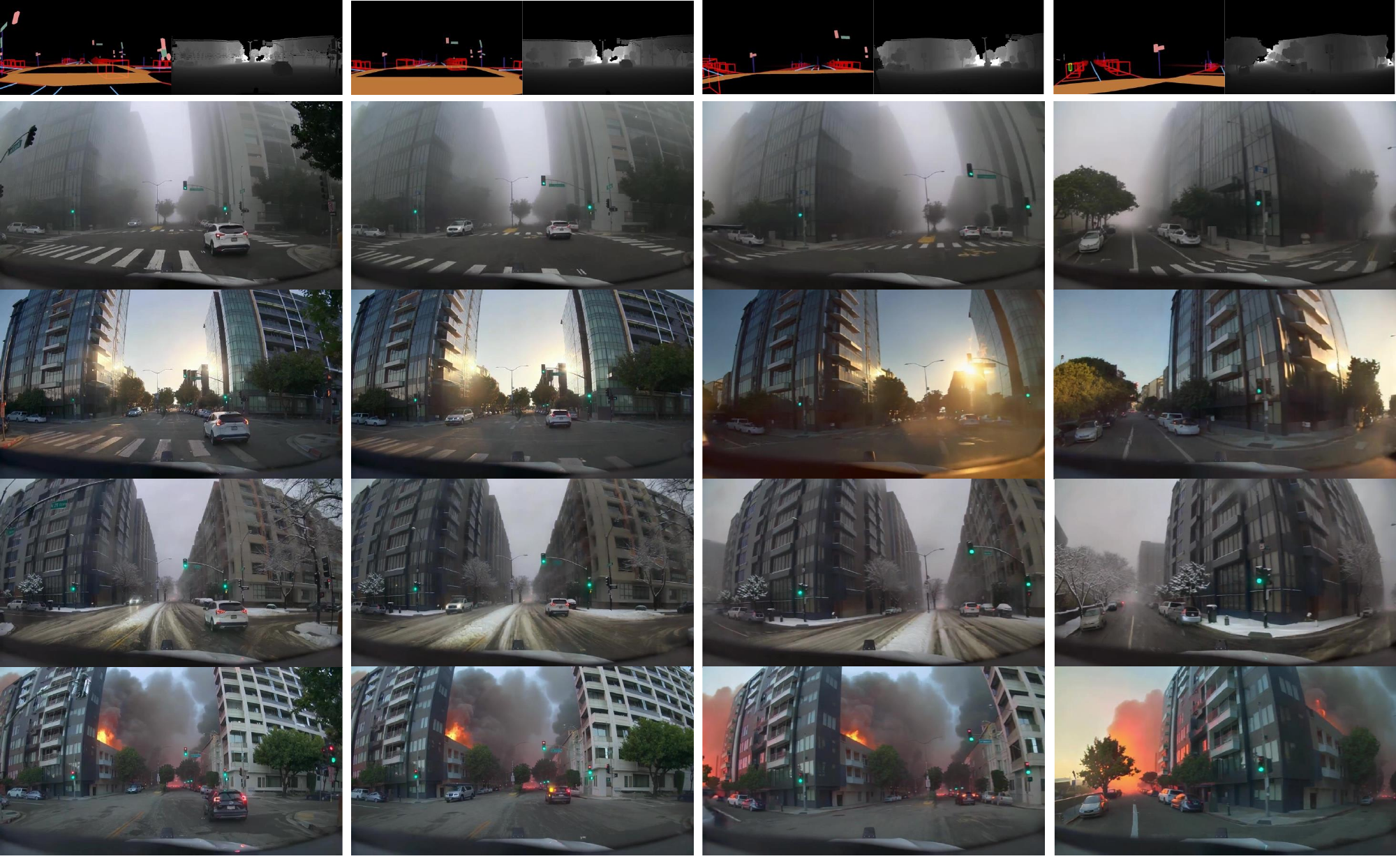}
    \caption{\textbf{1st row}: Control signals (left: HDMap + 3DBbox, right: LiDAR) to \sevenbav. \textbf{2nd-5th rows}: Video generated by different text prompts listed as following: \textit{The scene unfolds on a foggy morning, with a thick layer of mist reducing visibility...}; \textit{The scene is bathed in the warm, golden hues of the late afternoon sun, casting long shadows on the road...}; \textit{The street is blanketed in heavy snowfall, with large snowflakes continuously falling, partially obscuring visibility...}; \textit{The scene unfolds in a chaotic and intense environment as a fire engulfs the houses on either side of the street...}}
    \label{fig:driving_example}
\end{figure}

\cref{fig:av_depth_segment} presents the comparison between generation results using a single control signal with \sevenb---either Depth or Segmentation---and those obtained by combining both with \sevenb. When integrating multiple control signals in the examples of the figure, we simply adopt a uniform spatiotemporal control map, setting $w_\text{depth} = 0.5$ and $w_\text{segment} = 0.5$. In the first row, combining depth and segmentation restores the solid line in the middle of the road, which is however absent when using depth alone. This solid line is essential for creating a scenario where the ego vehicle must momentarily drive in the opposite lane to navigate around the safety zone delineated by traffic cones.
In the second row, segmentation alone generates unrealistic vehicle headings within the same lane, possibly due to the lack of detailed geometric cues. By integrating depth information, the fused depth and segmentation control signal produces more plausible vehicle orientations, improving the overall realism in the generated scene.

\cref{fig:driving_compare} similarly compares a single control signal of HDMap or LiDAR against their combined adaptive multimodal control signal based on \sevenbav. In these experiments, we combine HDMap and LiDAR control signals by setting uniform scalar values $w_\text{map} = 0.3$ and $w_\text{lidar} = 0.7$ for the spatiotemporal control map. In this example, the rightmost lane is blocked by traffic cones, and the ego vehicle attempts a right turn using the second right lane. In the fourth row, relying solely on LiDAR results in incorrect lane markings. In contrast, the fifth row demonstrates that fusing LiDAR with HDMap significantly enhances the overall realism of the generated lane layout. Additionally, \cref{fig:driving_example} presents more examples illustrating various generation results by different text prompts.

\begin{figure}[t!]
    \centering
    \includegraphics[width=\textwidth]{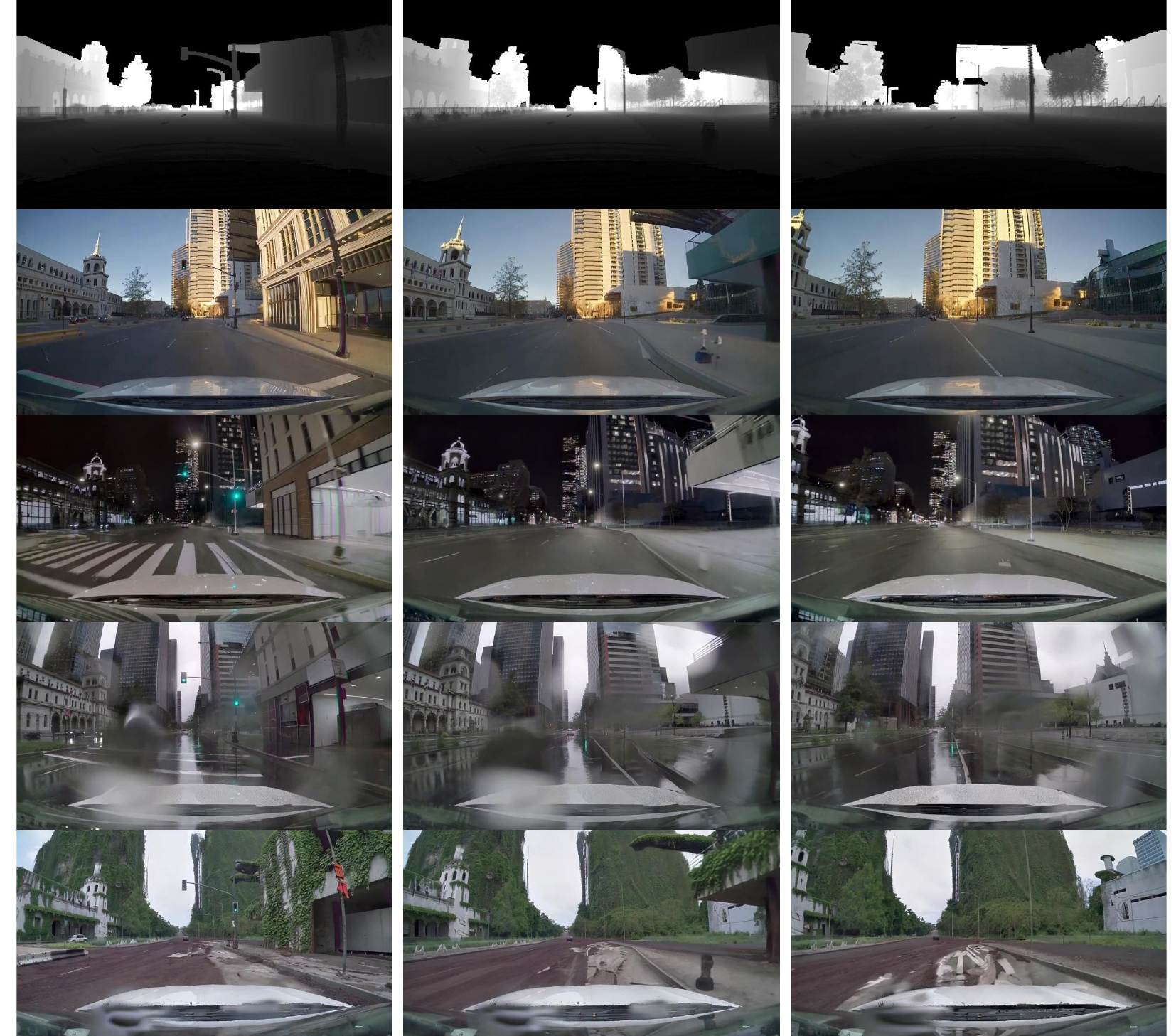}  
    \caption{\textbf{1st row}: LiDAR simulated by NVIDIA Omniverse as the control signal to \sevenb. \textbf{2nd-5th rows}: Videos generated by different text prompts listed as following: \textit{The video showcases an urban driving scene during the golden hour...}; \textit{The video portrays a nighttime driving scene in an urban environment...}; \textit{The video captures an urban driving scene under heavy rainfall...}; \textit{The video depicts a thrilling driving scene in a jungle-style urban environment...}}
    \label{fig:drivingsim}
\end{figure}

We further assess the generation quality of a single control signal (HDMap or LiDAR) and their combination using a diverse set of quantitative metrics in~\cref{tab:eval_av}. To assess adherence to the 3D object conditions, we utilize a 3D detection method built upon StreamPetr~\citep{streampetr} and Hydra-MDP~\citep{li2024hydra}. We compute the mean Average Precision (mAP) at an IoU threshold of $0.2$ by comparing the predicted 3D bounding boxes on \sevenbav-generated videos with their real-world counterparts. To measure adherence to the HDMap, we employ the grounded SAM2~\citep{ravi2024sam,liu2024grounding} model for lane segmentation and compute the IoU between the predicted lane areas and the actual lane markings. Lastly, to quantitatively assess the 3D consistency of generated video with respect to the conditioned LiDAR data, we evaluate the photometric reprojection error following~\citep{zhou2017unsupervised}. Specifically, we first subsample the generated video frames, retaining only those synchronized with the original LiDAR scans at 10 FPS. Using corresponding LiDAR-derived depths, known camera poses, and bounding box annotations, we warp each subsampled frame forward to the subsequent retained frame. We then compute the pixel-wise photometric discrepancy between each warped frame and the corresponding generated next frame using the L1 loss.

As shown in~\cref{tab:eval_av}, \sevenbav[LiDAR] achieves the best 3D-Bbox mAP scores, as it provides the most precise spatial conditioning signal. It also attains the lowest reprojection error, benefiting from LiDAR's inherent 3D awareness. However, it performs the worst in lane mIoU due to the absence of explicit lane layout information. In contrast, \sevenbav, which integrates both HDMap and LiDAR, achieves the highest lane mIoU while maintaining 3D-Bbox mAP, and its reprojection error score is comparable to \sevenbav[LiDAR] and better than \sevenbav[HDMap]. This highlights the benefits of integrating HDMap and LiDAR for improving scene generation, particularly in enhancing lane accuracy---one of the key metrics critical to AV developers.

Finally, we demonstrate \sevenbav for amplifying data variation in AV simulation. While previous examples rely on having a real-world counterpart of a scene layout, we can also connect \modelname with physically-based sensor simulation using the NVIDIA Omniverse Blueprint for AV simulation~\citep{avsim}. The blueprint is a reference workflow to create rich 3D worlds for AV training and testing. It contains APIs and services to model physics and behavior of dynamic objects in a scene and generate physically accurate and diverse sensor data.

In our experiment, we use Sensor RTX APIs in the blueprint to simulate LiDAR depth. As shown in~\cref{fig:drivingsim}, we observe that, despite being trained on real LiDAR, the model generalizes well to synthetic LiDAR. As a bonus, the model effectively retains world knowledge from internet pre-training, enabling generation of scenes and effects not captured in our AV fine-tuning dataset. This demonstrates that \modelname can become an effective tool to enhance the visual diversity of simulated scenes.

\begin{table}[t!]
    \centering
    \caption{\textbf{Quantitative evaluation of Cosmos-Transfer1-7B-Sample-AV on the autonomous driving video generation task.} We compare the results of both single-control models and multimodal control variant over various metrics. Best results are in bold.}
    \resizebox{\textwidth}{!}{
    \begin{tabular}{lccc}
        \toprule
         \multicolumn{1}{c}{Method} & 3D-Bbox mAP $\uparrow$  & Lane mIoU $\uparrow$ & Reprojection Err. $\downarrow$\\
        \midrule
        Cosmos-Transfer1-7B-Sample-AV [HDMap] & 41.89 & 50.37 & 9.46 \\
        Cosmos-Transfer1-7B-Sample-AV [LiDAR] & \textbf{46.50} & 48.19 & \textbf{8.60} \\
        Cosmos-Transfer1-7B-Sample-AV & 44.66 &\textbf{51.55} & 8.67 \\

        \bottomrule
    \end{tabular}
    }
    \label{tab:eval_av}
\end{table}

\section{Real-time Inference} \label{sec:inference}

In this section, we report an implementation of \sevenb that achieves real-time inference performance by leveraging the new NVIDIA GB200 NVL72 system. 
\textbf{GB200 NVL72} binds together 36 \textbf{G}race CPUs and \textbf{72} \textbf{B}lackwell GPUs, in an any-to-any \textbf{NVL}ink network. This architecture is ideal for model parallelism techniques, such as Tensor and Context Parallelism. These techniques are used by many large-scale foundation models~\citep{grattafiori2024llama}, including World Foundation Models~\citep{parkerholder2024genie2,nvidia2025cosmos}, in both training and inference scenarios. However, unlike Large Language Models (LLMs), which are often heavy in number of parameters and generate one token at a time, \sevenb is relatively lightweight in number of parameters and generates tens of thousands of tokens in one shot.

Considering this use case, we devise a parallelism strategy for scaling \sevenb in which we employ pure data parallelism in non-attention layers and head-parallelism in attention layers. Each B200 GPU packs up to 192GB of High-Bandwidth Memory (HBM), which can easily store an entire copy of the \sevenb model. Therefore, to generate a 5-second 720p video, we can shard the entire 56K token sequence among GPUs. Under this setting, the only point of communication during diffusion is in attention, where we use the all-to-all collective so that each GPU operates on the entire 56K token sequence from a single attention head. This approach is favorable to the pipelined all-gathering of the key-value pair, since it does not require additional reduction steps and does not result in heavily memory-bandwidth-bound attention. \sevenb has 32 attention heads, and uses classifier free guidance where the unconditional denoising is replaced with negative prompt-based denoising. We split the workload of the positive and negative conditionings among two groups of GPUs, as their denoising process is independent. As a result, we can distribute the attention workload among 64 GPUs so that each GPU performs attention over 56K query, key, and value tokens. Even with the largest query tile size used in Blackwell FMHA kernels~\citep{thakkar2023cutlass}, this still ensures full occupancy of the streaming multiprocessors (SMs) available in each B200 GPU.

\begin{table}[t!]
    \centering
    \caption{\textbf{Computation time for generating a 5-second video with \sevenb under different parallelism settings.} End-to-end runtime dips below 5 seconds when scaled up to 64 B200 GPUs and reach real-time generation throughput.}
    \begin{tabular}{ccccccc}
        \toprule
        Number of GPUs           & 1       & 4      & 8      & 16     & 32    & 64     \\
        \midrule
        Diffusion only           & 141.0 s & 39.3 s & 20.1 s & 10.3 s & 5.4 s & 3.5 s  \\
        End-to-end               & 141.7 s & 40.0 s & 20.8 s & 11.0 s & 6.1 s & 4.2 s  \\
        \bottomrule
    \end{tabular}
    \label{tab:gb200_inference}
\end{table}

We report generation times when using different numbers of GPUs in \cref{tab:gb200_inference}. As shown, our parallelism strategy exhibits an approximately 40X speedup when going from 1 to 64 GPUs, when we only consider diffusion runtime, which is over 99\% of the workload, and the only workload we parallelize across GPUs. We also achieve real-time generation throughput when using 64 GPUs, generating 5 seconds of video in only 4.2 seconds.

\section{Related Work} \label{sec::related}

\noindent{\bf Visual Domain Transfer.} Numerous studies have explored the transfer of visual domains from abstract representations to photorealistic visualizations. A prominent line of research focuses on converting segmentation maps or sketches into high-fidelity images~\citep{wang2018pix2pixHD,park2019SPADE,dundar2020domain,sushko2020you,huang2022poegan,wang2021image,wang2022semantic,shi2022retrieval,lv2022semantic,xue2023freestyle,lv2024place,fontanini2025semantic}. Beyond static images, several works extend this paradigm to video synthesis~\citep{wang2018vid2vid,wang2019fewshotvid2vid,mallya2020world,zhuo2022fast,chung2023shortcut,esser2023structure,zhuo2024fast}, significantly enhancing the capability for dynamic and temporally coherent visual generation. These advancements have broad implications across various domains, including content creation, robotics, autonomous driving, virtual and augmented reality (VR/AR), and gaming.

\noindent{\bf Spatial Control for Diffusion Models.} Diffusion models have demonstrated remarkable capabilities in text-to-image and text-to-video generation. To enhance their spatial controllability, various methods have been proposed, which can be broadly categorized into training-free approaches \citep{xue2023freestyle,chen2023trainingfree,bansal2023universal} and techniques requiring additional training on top of pre-trained text-to-image models~\citep{huang2023composer,mou2024t2i,zhang2023adding,zhao2024uni,qin2023unicontrol,li2023gligen,liu2024humangaussian,ju2023humansd,liu2023hyperhuman,ren2025gen3c,zeng2023scenecomposer,lu2024infinicube}. Generally, the latter category achieves superior results. A notable work is ControlNet \citep{zhang2023adding}, which introduced an additional encoder branch initialized from a pre-trained model, updating only this branch during training. Following its success, several extensions have been proposed~\citep{zhao2024uni,qin2023unicontrol,ju2023humansd, sun2024anycontrol}. More recently, \cite{chen2024pixart} extended the applicability of ControlNet from UNet-based architectures to transformers. Spatial control has also been explored in video generation. \cite{lin2024ctrl} proposed to adapt a pre-trained ControlNet encoder from image to video. \cite{jain2024peekaboo} introduced a training-free approach leveraging masked attention modules to enable spatial control in video synthesis.

\noindent{\bf Enhancing Simulation with Generative Models.} Developing and testing Physical AI systems in real-world environments pose significant risks, making simulation a crucial component for these tasks. Recent advancements in generative AI have greatly improved simulation by enhancing its realism, diversity, and utility.Early works leveraged generative models to refine simulator outputs. \cite{rao2020rl} trained a CycleGAN to transform simulator-rendered images into more photorealistic counterparts while ensuring their utility by aligning the Q-values of the simulator and GAN-enhanced outputs. \cite{ho2021retinagan} proposed RetinaGAN, an unsupervised GAN framework that enhances thr realisim of simulated scenes while preserving essential object features, demonstrating effectiveness in reinforcement learning across multiple real-world tasks. More recently, diffusion models have emerged as a powerful alternative to GANs for simulation enhancement. \cite{zhao2024exploring} utilized a diffusion-based model with ControlNet for sim-to-real transfer, converting simulation outputs (\eg semantic maps) into photorealistic driving images. Their approach outperformed GAN-based methods, exhibiting fewer artifacts and better structural fidelity. \cite{pronovost2023scenario} introduced a scenario generation pipeline based on latent diffusion, synthesizing complex traffic environments to enable scalable testing of autonomous agents under diverse and safety-critical driving conditions. Beyond static image synthesis, video generation models are increasingly being explored as learnable simulators for planning and control. \cite{nvidia2025cosmos} presented a generative world model platform, providing a collection of open-source pre-trained models tailored for various physical AI tasks, further advancing simulation capabilities for real-world applications.

\section{Conclusion} \label{sec::conclusion}

We introduced \modelname, a diffusion-based conditional world model for multimodal controllable world generation. By introducing multimodal control branches to Cosmos-Predict1~\citep{nvidia2025cosmos} with an adaptive weighting scheme, \modelname enables highly controllable world generation and improves generation quality. Extensive evaluations demonstrate that \modelname effectively preserves scene structure from the condition inputs while allowing fine-grained control, making it a powerful tool for bridging the synthetic-to-real domain gap in applications such as robotics Sim2Real and autonomous vehicle data enrichment. Additionally, our inference scaling strategy enables real-time throughput generation with an NVIDIA GB200 NVL72 rack, highlighting its practical feasibility. We open-sourced our code and models to help advance Physical AI research.

\clearpage
\appendix
\section{Prompt Upsampler}
\label{sec:prompt_upsampler}

One of the primary objectives of \modelname is to accommodate a diverse range of user requests, which can vary significantly in style, format, and content.
During training, \modelname generates videos based on multimodal inputs and detailed descriptions. The detailed descriptions may differ substantially from user queries, leading to suboptimal performance. To mitigate this domain shift, we train a prompt upsampler to align diverse input prompts with the training prompt distribution of \modelname.

Following \citet{nvidia2025cosmos}, we aim to develop prompt upsamplers that satisfy the following three key criteria
\begin{itemize}
    \item \textbf{Fidelity.} The upsampled prompt should remain faithful to the original short prompt and its corresponding conditioned video, avoiding any semantic conflicts.

    \item  \textbf{Completeness.} The expanded prompt should fully retain and elaborate upon all elements present in the original prompt.

    \item 
 \textbf{In-Distribution Consistency.} The structure and length of the upsampled prompt should align with the distribution of long prompts in the training dataset, ensuring optimal performance.
\end{itemize} 

Specifically, we finetune one Pixtral-12B \citep{agrawal2024pixtral} to take in both the condition video and a user prompt, transforming it into a more detailed prompt with consistent structure.
The input condition video can be of different modalities (\emph{e.g.} segmentation mask, depth). 
To curate high-quality training data for this task, we begin with existing training prompts and utilize \texttt{Gemma-2-9B-it} \citep{team2024gemma} to generate diverse short prompts that preserve fidelity while exhibiting varied characteristics.
We curate a paired dataset of 1M videos for each modality and jointly train on all modality videos for one epoch with FSDP2 \citep{zhao2023pytorch}.

For example, the prompt upsampler can understand and recognize key elements from segmentation maps and transform input text with rich information:

\begin{tcolorbox}[colback=nvidiagreen!3,colframe=nvidiagreen!75!white,title=\textsc{Example output from upsampler that takes in short prompt and a segmentation map, and transform into longer prompts with details.},left=0.5ex,right=0.5ex,top=0.5ex,bottom=0.5ex]
\fontsize{9.}{9.}\selectfont
\begin{minipage}[t]{0.45\textwidth}
\begin{Verbatim}[breaklines=true, breaksymbolleft={}, breaksymbolright={}]
A robot in the kitchen picks up a bottle from the floor and puts it on a table.
\end{Verbatim}
\includegraphics[width=\linewidth]{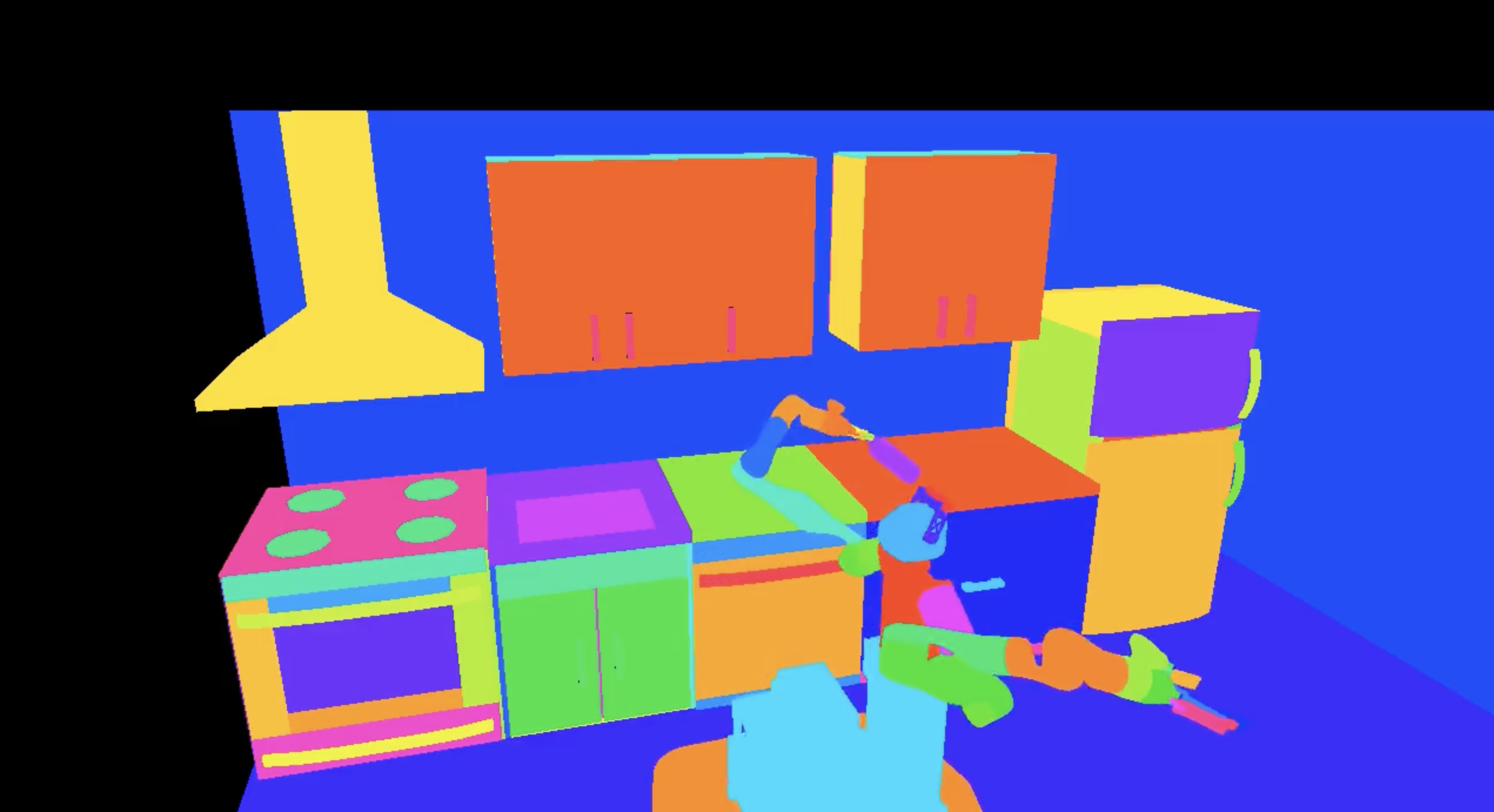}
\end{minipage} 
\hfill
\begin{minipage}[t]{0.45\textwidth}
\begin{center}
\textbf{\underline{OUTPUT}}
\end{center}
\begin{Verbatim}[breaklines=true, breaksymbolleft={}, breaksymbolright={}]
The video features a kitchen with wooden cabinets and a granite countertop. A robot with a white body, black joints, and a red light on its head is seen performing tasks. It moves its arms and legs to pick up a white bottle with a red label from the floor and place it on the countertop. The robot then moves to a dining area with a wooden table and chairs, where it picks up a white chair and places it back in its original position.
\end{Verbatim}
\end{minipage}
\end{tcolorbox}

In a similar spirit, for depth input, the prompt upsampler also develops a holistic understanding of the scene and can recognize key surrounding objects:

\begin{tcolorbox}[colback=nvidiagreen!3,colframe=nvidiagreen!75!white,title=\textsc{Example output from upsampler that takes in short prompt and depth, and transform into longer prompts with details.},left=0.5ex,right=0.5ex,top=0.5ex,bottom=0.5ex]
\fontsize{9.}{9.}\selectfont
\begin{minipage}[t]{0.45\textwidth}
\begin{Verbatim}[breaklines=true, breaksymbolleft={}, breaksymbolright={}]
A car is driving on a road.
\end{Verbatim}
\begin{center}
    \includegraphics[width=\linewidth]{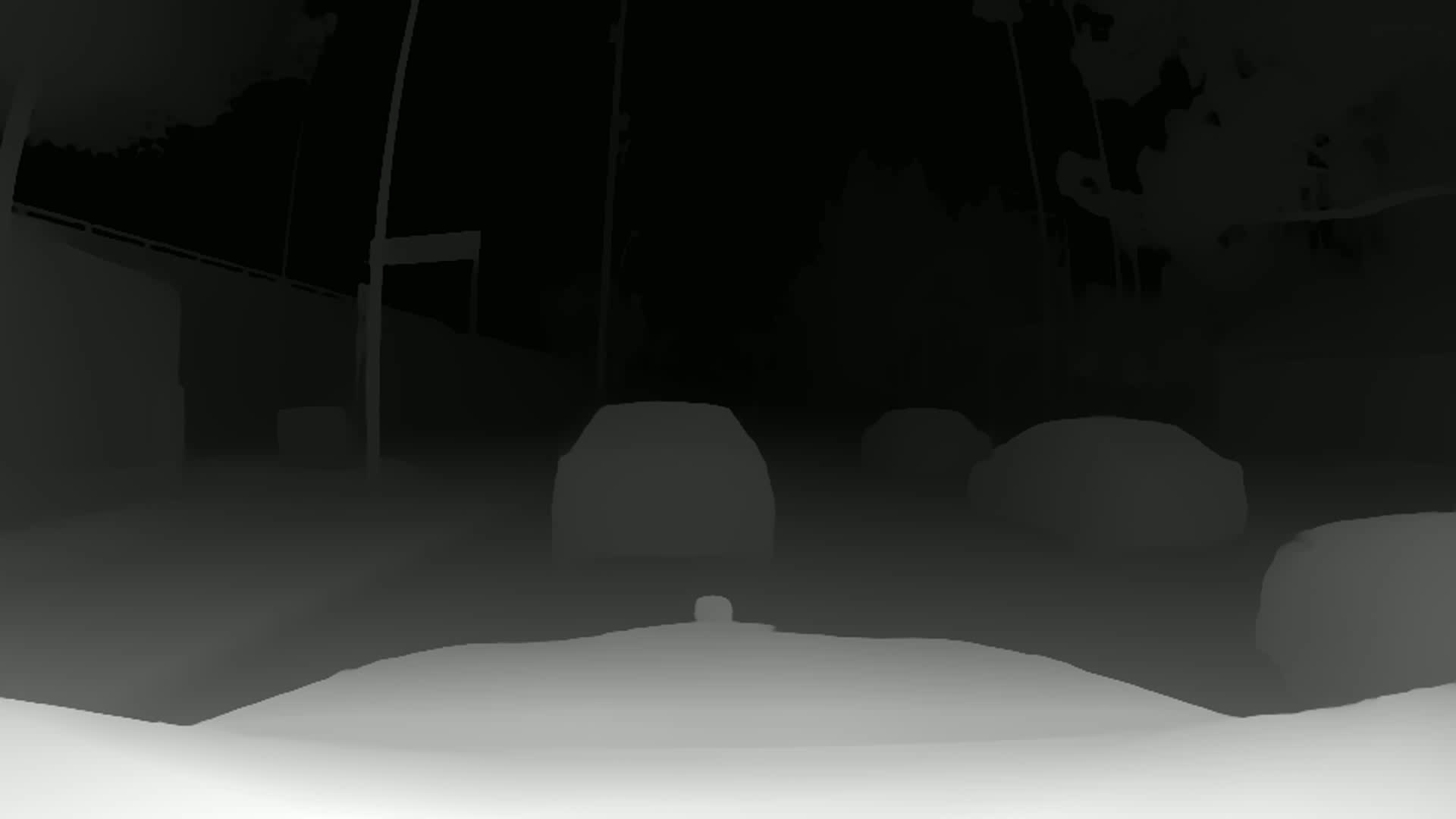}
\end{center}
\end{minipage} 
\hfill
\begin{minipage}[t]{0.45\textwidth}
\begin{center}
\textbf{\underline{OUTPUT}}
\end{center}
\begin{Verbatim}[breaklines=true, breaksymbolleft={}, breaksymbolright={}]
The video is taken from the perspective of a car's dashboard, showing the view of the road ahead. It is daytime with clear skies. The road is surrounded by greenery, with trees and shrubs visible on both sides. The car is driving on a road with multiple lanes, and there are other vehicles visible in the video, including a motorcycle. The lighting suggests it is daytime, and the weather appears to be clear.	
\end{Verbatim}
\end{minipage}
\end{tcolorbox}
\clearpage
\section{Contributors and Acknowledgements}
\label{sec:contributors}

\subsection{Core Contributors}
Maciej Bala, Tiffany Cai, Francesco Ferroni, Sanja Fidler, Dieter Fox, Yunhao Ge, Jinwei Gu, Ali Hassani, Pooya Jannaty, Huan Ling, Ming-Yu Liu, Xian Liu, Yifan Lu, Qianli Ma, Hanzi Mao, Fabio Ramos, Xuanchi Ren, Tianchang Shen, Shitao Tang, Ting-Chun Wang, Jiashu Xu, Xiaodong Yang, Xiaohui Zeng, Yu Zeng

\noindent \textbf{Contributions:} 
\textbf{MYL} initiated the adaptive Multimodal Control design. \textbf{TCW} and \textbf{XZ} trained the individual ControlNets of Cosmos-Transfer1-7B. \textbf{TCW}, \textbf{TC}, \textbf{PJ} implemented the Multimodal Control inference. \textbf{ST}, \textbf{YG}, \textbf{QM}, \textbf{HM} contributed to the Cosmos-Transfer1-7B training data curation. \textbf{FR}, \textbf{DF}, \textbf{JG} and \textbf{YG} curated robotics simulation data. \textbf{YL}, \textbf{XR}, \textbf{TS} curated the RDS-HQ dataset. \textbf{QM}, \textbf{HM}, \textbf{MYL} designed the evaluation framework. \textbf{FF} and \textbf{QM} built the evaluation benchmark and framework. \textbf{QM}, \textbf{FF}, \textbf{JG}, \textbf{YG}, \textbf{XY}, \textbf{HM} conducted evaluations for Cosmos-Transfer1-7B. \textbf{XR}, \textbf{TS}, \textbf{HL} trained the individual ControlNets of Cosmos-Transfer1-7B-Sample-AV and conducted corresponding experiments. \textbf{JX} and \textbf{YG} built the prompt upsampler. \textbf{AH} and \textbf{MB} designed and implemented real-time inference with the NVIDIA GB200 NVL72 system. \textbf{SF} and \textbf{HL} supervised the Cosmos-Transfer1-7B-Sample-AV research, data curation, model training and evaluation. \textbf{HM} supervised the Cosmos-Transfer1-7B research, data curation, model training and evaluation. 
All core authors contributed to paper writing. \textbf{HM} and \textbf{MYL} organized paper writing. \textbf{MYL} supervised the project.

\subsection{Contributors}
Hassan Abu Alhaija, Jose Alvarez, Tianshi Cao, Liz Cha, Joshua Chen, Mike Chen, Michael Isaev, Shiyi Lan, Tobias Lasser, Alice Luo, Xinglong Sun, Jay Wu, Kevin Xie, Stella Xu, Yuchong Ye

\subsection{Acknowledgments}
Joshua Bapst, Prithvijit Chattopadhyay, TJ Galda, Mingfei Guo, Madison Huang, Hai Loc Lu

\clearpage
\setcitestyle{numbers}
\bibliographystyle{plainnat}
\bibliography{main}

\end{document}